\documentclass[lettersize,journal]{IEEEtran}
\usepackage{amsmath,amsfonts}
\usepackage{bm}
\usepackage{algorithmic}
\usepackage{array}
\usepackage[caption=false,font=footnotesize,labelfont=sf,textfont=sf]{subfig}
\usepackage{textcomp}
\usepackage{stfloats}
\usepackage{url}
\usepackage{verbatim}
\usepackage{graphicx}
\def\BibTeX{{\rm B\kern-.05em{\sc i\kern-.025em b}\kern-.08em
    T\kern-.1667em\lower.7ex\hbox{E}\kern-.125emX}}
\usepackage{balance}

\usepackage{float}
\usepackage{booktabs}
\usepackage{multirow}

\usepackage{tabularx} 

\begin{document}
\title{SPLIT: Sparse Incremental Learning of Error Dynamics for Control-Oriented Modeling in Autonomous Vehicles}
\author{Yaoyu Li, Chaosheng Huang$^{\ast}$, and Jun Li%
\thanks{$^{\ast}$Corresponding author: Chaosheng Huang.}%
\thanks{Yaoyu Li, Chaosheng Huang and Jun Li are with School of Vehicle and Mobility, Tsinghua University, Beijing 100084, China.
(e-mail: liyy20@mails.tsinghua.edu.cn; huangchaosheng@tsinghua.edu.cn; lijun1958@tsinghua.edu.cn).}%
}
\markboth{Journal of \LaTeX\ Class Files,~Vol.~18, No.~9, September~2020}%
{How to Use the IEEEtran \LaTeX \ Templates}

\maketitle

\begin{abstract}

Accurate, computationally efficient, and adaptive vehicle models are essential for autonomous vehicle control. 
Hybrid models that combine a nominal model with a Gaussian Process (GP)-based residual model have emerged as a promising approach.
However, the GP-based residual model suffers from the curse of dimensionality, high evaluation complexity, and the inefficiency of online learning, which impede the deployment in real-time vehicle controllers.
To address these challenges, we propose SPLIT, a sparse incremental learning framework for control-oriented vehicle dynamics modeling. SPLIT integrates three key innovations: (i) Model Decomposition. We decompose the vehicle model into invariant elements calibrated by experiments, and variant elements compensated by the residual model to reduce feature dimensionality. (ii) Local Incremental Learning. We define the valid region in the feature space and partition it into subregions, enabling efficient online learning from streaming data.
(iii) GP Sparsification. We use bayesian committee machine to ensure scalable online evaluation. Integrated into model-based controllers, SPLIT is evaluated in aggressive simulations and real-vehicle experiments. 
Results demonstrate that SPLIT improves model accuracy and control performance online. Moreover, it enables rapid adaptation to vehicle dynamics deviations and exhibits robust generalization to previously unseen scenarios.

\end{abstract}


\section{Introduction}
\label{section: introduction}

\IEEEPARstart{M}{odel-based} controllers have been widely employed in autonomous vehicles as an attractive framework for achieving safety and performance \cite{li2025learning, askari2025Model, stano2023model, hewing2020learning}. Vehicle dynamics models constitute the cornerstone of model-based controllers, providing predictive representations of vehicle dynamics that enable the generation of precise control inputs \cite{chen2020implementation,armin2023Integrating}. To serve this role effectively, vehicle dynamics models must satisfy three stringent requirements: fidelity, to capture essential and complex vehicle dynamics; computational efficiency, to enable execution within real-time control cycles; and adaptability, to account for variations in vehicle dynamics and environmental conditions. Nevertheless, simultaneously meeting these requirements is notoriously difficult: high-fidelity models usually entail prohibitive computational complexity, while variations in vehicle dynamics are inherently unpredictable and difficult to observe directly. Constructing control-oriented models thus remains a longstanding challenge in autonomous vehicles \cite{yang2013overview,zhang2024survey}.

Among existing practices for control-oriented modeling, the integration of a physical model with a Gaussian Process (GP)-based residual model constitutes a highly promising modeling paradigm with the potential to satisfy the stringent requirements \cite{hewing2018cautious,kabzan2019learning,hewing2019cautious,jiang2021high}. The physical model preserves interpretable structure and delivers stable baseline predictions. The GP-based residual model then compensates for analytically intractable discrepancies to refine accuracy where the physical model alone falls short.
This hybrid framework benefits from the probabilistic nature of GP inference, which yields interpretable predictive distributions. In addition, GP predictions naturally revert to the prior mean in regions insufficiently covered by training set, preventing unwarranted corrections to the physical baseline. Moreover, the GP formulation admits incremental updates, offering a pathway to online adaptation without the need for exhaustive retraining.

\begin{figure*}[!t]
\centering
\includegraphics{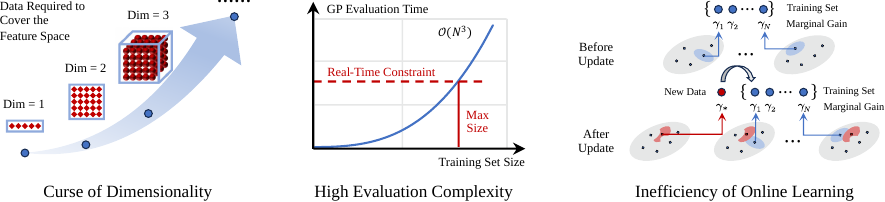}
\caption{
Challenges of the hybrid modeling paradigm integrating a physical model and a GP-based residual model.
(a) Curse of dimensionality. Red circles denote data, while blue lines delineate unit volumes in different feature space dimensions. Accurate GP predictions require sufficient coverage of the feature space, yet the amount of data needed grows exponentially with dimensionality, making dense coverage of high-dimensional vehicle dynamics feature space practically intractable.
(b) High evaluation complexity. GP evaluation necessitates computational complexity of $\mathcal{O}(N^3)$. Real-time constraints in controllers limit the feasible training set size, which narrows the applicable operating range of the model and compromises model accuracy.
(c) Inefficiency of online learning. Red and blue circles represent new and existing data, while light ellipses denote the marginal gain of each data for feature space coverage. Online learning of GPs requires evaluating the marginal gain of new data and re-evaluating that of existing data, which slows down the learning process and hinders timely adaptation to model deviations.
}
\label{figure: challenges}
\end{figure*}

However, despite its promising capability, this hybrid modeling paradigm still encounters three fundamental challenges when considered for deployment in real-time vehicle control systems, as illustrated in Fig. \ref{figure: challenges}:

\begin{enumerate}
\item \textbf{Curse of Dimensionality}. Capturing complex model discrepancies typically requires high-dimensional feature representations, which cause the volume of the feature space to grow exponentially. Consequently, acquiring data with sufficient density to ensure adequate coverage becomes practically intractable in real-world.
\item \textbf{High Evaluation Complexity}. 
Performing evaluation with GPs necessitates the construction and inversion of a kernel matrix over the entire training set, leading to a computational complexity of $\mathcal{O}(N^3)$ with respect to the training set size. Such scaling quickly becomes prohibitive under the millisecond-level sampling intervals of real-time vehicle controllers.
\item \textbf{Inefficiency of Online Learning}. Online GP learning sequentially updates the training set by evaluating the marginal contribution of each new sample to the model representation. Each update entails evaluating the incoming sample and re-evaluating the entire training set, which makes the process prohibitively slow and prevents timely adaptation to evolving vehicle dynamics
\end{enumerate}
Therefore, existing research efforts are predominantly constrained to fixed driving scenarios, such as race tracks, where small-scale training sets are constructed and updated offline, yielding only partial error compensation \cite{kabzan2019learning, ostafew2014learning, ostafew2016robust}. Consequently, the practical applicability of GP-based residual models remains severely limited.

To systematically address the three aforementioned challenges, this article substantially extends the recent work \cite{li2025learning} into an online learning formulation and proposes SPLIT—a sparse incremental learning framework. The central principle of SPLIT is to leverage physically motivated assumptions and experimentally verifiable calibration to reduce both the dimensionality of the learning problem, and the computational burden of online learning and evaluation. 
SPLIT strategically sacrifices a marginal degree of model fidelity in exchange for real-time feasibility and rapid adaptability, enabling efficient online learning and error compensation across the entire performance envelope.
Overall, SPLIT operates as an integrated vehicle dynamics modeling framework that aligns more closely with the requirements of model-based control.

The primary contributions of this article are summarized as follows.

\begin{enumerate}
\item We decompose the vehicle model into invariant elements calibrated through experiments, and variant elements compensated by a GP-based residual model. This decomposition reduces the feature dimensionality of the residual model and exponentially decreases the required training set size. 
\item We explicitly define the valid region within the feature space and partition it into equally sized cubic subregions. This partitioning transforms the challenge of updating the training set across the entire performance envelope into a manageable task of updating individual subsets, which supports online learning.
\item Based on the partitioned data subsets, we employ the Bayesian Committee Machine (BCM) to sparsify the online evaluation of the GP, substantially reducing its computational complexity.
\item SPLIT is integrated into model-based controllers and validated through both simulations and real-world experiments under various aggressive scenarios. The experimental results demonstrate that SPLIT can continuously refine model accuracy online and improve closed-loop control performance. Moreover, SPLIT exhibits rapid adaptation to model deviations and robust generalization to previously unseen scenarios.
\end{enumerate}

The rest of the paper is organized as follows: 
Section \ref{section: related work} reviews related works.
Section \ref{section: the vehicle model} outlines the vehicle model. 
In Section \ref{section: sparse incremental learning of error dynamics}, sparse incremental learning mechanism of error dynamics is presented. 
Section \ref{section: model-based controller formulation} formulates the model-based controllers.
Section \ref{section: simulation experiments} reports simulation experiments, and Section \ref{section: real-world experiments} presents real-world experiments. Finally, Section \ref{section: conclusion} concludes the paper.


\section{Related Work}
\label{section: related work}
Numerous control-oriented vehicle dynamics modeling approaches have been proposed in an effort to meet the stringent requirements of model-based controllers \cite{yang2013overview,zhang2024survey}. Broadly, these approaches can be categorized into three types: physical model, data-driven model, and hybrid model. A concise review is presented below.

\subsection{Physical Model}
\label{subsection: physical model}

Physical model is established from first principles to formulate differential equations that mathematically represent vehicle dynamics \cite{kapania2015design, brown2017safe}. 
Attributable to their interpretability, they constitute the most extensively used class of models \cite{subosits2019racetrack, falcone2007predictive, brown2019coordinating, christ2021time}.
However, improving the fidelity of such models requires incorporating detailed descriptions of the vehicle system, which inevitably increases the state dimensionality and model complexity. Moreover, physical models struggle to accurately capture complex dynamics that lack precise analytical representations, especially tire–road interactions, which remain a dominant source of modeling error \cite{guo2018tire, xu2024combined}. More critically, these models rely on parameters obtained through offline calibration, leaving them inflexible to dynamics variations \cite{kabzan2019learning, spielberg2019neural, spielberg2021neural}. To alleviate these limitations, a variety of adaptive control approaches have been developed for online parameter estimation and have been applied in vehicle control \cite{wenzel2006dual, qin2022lateral, wang2021integrated, li2021dual, zhang2023tire}. Nevertheless, these methods often suffer from overfitting to recent observations and lack guarantees of convergence to true parameters \cite{nguyen2011model,brunke2022safe}. And the parameter estimation grows more challenging as the model complexity increases.

\subsection{Data-Driven Model}
\label{subsection: Data-Driven Model}

Data-driven model has also been employed to address the limitations of physical models by directly reconstructing the mapping of vehicle dynamics from empirical driving data \cite{hewing2020learning, armin2023Integrating, kuutti2020survey}. 
Their primary advantage lies in the superior capacity to capture complex dynamics, especially under extreme operating conditions where physical models often fail to provide accurate descriptions.
In \cite{spielberg2019neural} and \cite{spielberg2021neural}, neural network vehicle models are employed to enable vehicle control under conditions close to the tire–road adhesion limits. In \cite{djeumou2023autonomous, ding2024drifting, djeumou2024one} and \cite{suminaka2025adaptable}, data-driven vehicle models are employed for more aggressive drifting conditions beyond the tire–road adhesion limits.
Moreover, when vehicle dynamics vary, data-driven models can implicitly recognize such variations through exposure to diverse training data \cite{spielberg2019neural, spielberg2021neural, kim2022toast}, and can further adapt by leveraging real-time measurements for online learning \cite{ding2024drifting, rutherford2010modelling, williams2017information, davydov2025first}, which enables adaptability to dynamics variation.

Nevertheless, the non-transparent nature of data-driven models makes it difficult to ensure the interpretability of their outputs. And their generalization capability to unseen operating conditions is also limited. 
Hence, the latest research has attempted to incorporate physical priors into learning frameworks, including pre-training with data generated from physics-based models \cite{spielberg2019neural, spielberg2021neural}, augmenting training sets according to physical laws \cite{ding2024drifting, davydov2025first}, embedding physical structures into neural networks \cite{djeumou2024one, kim2022physics, chrosniak2024deep}, or encoding physical models into loss functions \cite{cao2024intelligent, yang2024trajectory}. Yet, these approaches are still not sufficient to fundamentally overcome the inherent shortcomings. Furthermore, online learning often suffers from catastrophic forgetting, preventing consistent retention and integration of knowledge across heterogeneous scenarios.

\subsection{Hybrid Model}
\label{subsection: hybrid model}
Hybrid model typically retains a simple nominal model to capture essential dynamics, while augmenting it with a GP-based residual model to capture unmodeled dynamics, leveraging the complementary strengths of physical and data-driven models \cite{mesbah2022fusion, scampicchio2025gaussian}.
To balance the trade-off between feature space coverage and the computational complexity of online GP evaluation, existing hybrid models typically employ small-scale training sets to provide only local coverage of the feature space. 
In \cite{ostafew2014learning, ostafew2016robust, ostafew2016learning}, data subsets are constructed based on path vertices and velocities for training set, and the resulting GP is then used to reduce trajectory tracking errors in off-road scenarios. In \cite{hewing2018cautious, kabzan2019learning, hewing2019cautious, kabzan2020amz}, data dictionaries of limited size are constructed for GP, which is subsequently employed to improve lap time performance in autonomous racing. This strategy can improve model accuracy within limited operating conditions but performs poorly in unseen scenarios, as different operating conditions correspond to different regions of the feature space. 
\cite{ostafew2016robust, ostafew2016learning, mckinnon2017learning, jiang2023fast, zhao2024learning} construct multiple candidate subsets offline for online switching, but in high-dimensional feature spaces these approaches have limited efficacy and offer only marginal improvements in generalization.

Moreover, sparsification of GPs can reduce the computational cost of online evaluation, enabling the larger training sets. Common methods include low-rank approximations \cite{kabzan2019learning, hewing2019cautious, snelson2005sparse}, variational inference \cite{jiang2021high, hensman2015scalable}, and sparse spectrum GPs \cite{lazaro2010sparse, pan2017prediction}. Yet when training sets are sufficiently large to cover the entire high-dimensional feature space, no single method can reduce the computational time to a level acceptable for real-time control. Therefore, fundamentally reducing the dependence on training set size constitutes an indispensable step for the practical deployment of hybrid models.

Adaptation to vehicle dynamics variations relies on updating the training set with streaming data. 
\cite{ostafew2014learning, ostafew2016robust, ostafew2016learning} and \cite{mckinnon2017learning} retain the most recent measurements to enable real-time adaptability, but such strategies introduce substantial redundancy.
\cite{kabzan2019learning} proposes a more principled approach by quantifying the marginal contribution of each data point to feature space coverage as a criterion for sample selection, yet this evaluation becomes computationally prohibitive as the training set size grows.
Consequently, such processes are often executed offline \cite{jiang2023fast, zhao2024learning} or in threads with relaxed real-time requirements \cite{kabzan2019learning, kabzan2020amz}, severely constraining the ability of the model to rapidly adapt to changing vehicle dynamics.

In contrast to existing research, the proposed SPLIT framework extends beyond the local coverage paradigm and refines model accuracy online across the full performance envelope.

For the modeling framework, we retain the standard hybrid model framework \cite{hewing2018cautious, kabzan2019learning, hewing2019cautious}, but restrict the residual model to compensateing only for variant elements deviations. This decomposition reduces the feature dimensionality of the residual model and exponentially decreases the required training set size.

For updating the GP training set, we adopt the mechanism of \cite{kabzan2019learning, nguyen2011incremental}, but apply it only to the subsets corresponding to streaming data to enable online updates. 
The similar idea is also considered in \cite{jiang2023fast, zhao2024learning}, yet the resulting subsets do not guarantee coverage of the feature space.
In contrast, SPLIT generates data subsets by explicitly defining and partitioning valid regions, ensuring uniform coverage across the vast majority of driving conditions.

For GP online evaluation, we avoid the full kernel matrix re-inversion required after each training set update in \cite{pan2017prediction,jiang2021high}, and the prediction oscillations caused by dynamic selection of inducing points in \cite{hewing2018cautious, kabzan2019learning, hewing2019cautious}. Instead, training set updates in SPLIT modify only the kernel matrices of local subsets, and predictions are produced by constructing GPs on the subsets and aggregating them via BCM. Furthermore, the parallelization of multiple GPs further reduces the evaluation time to a minimal level.

\section{The Vehicle Model}
\label{section: the vehicle model}

\begin{figure}[!t]
\centering
\includegraphics{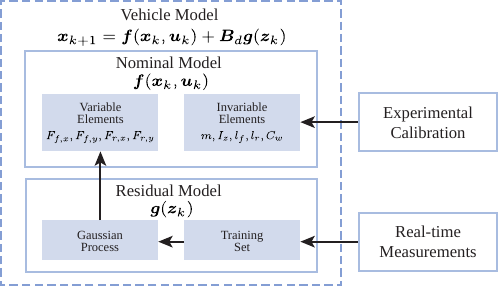}
\caption{Architectural of the vehicle model. The vehicle model consists of a nominal model and a residual model. The parameters \( m \), \( I_z \), \( l_f \), \( l_r \), and \( C_d \) in nominal model are defined as invariable elements, as their variations are negligible and can be determined through calibration. Conversely, the tire forces \( F_{f,x} \), \( F_{f,y} \), \( F_{r,x} \), and \( F_{r,y} \) are defined as variable elements due to their potential fluctuations caused by factors such as temperature, tire pressure, and road surface conditions, which are compensated by the residual model.}
\label{figure: vehicle model framework}
\end{figure}


In this section, we introduce the framework of the control-oriented vehicle dynamics model, as illustrated in Fig. \ref{figure: vehicle model framework}, including both the nominal physical model in Section \ref{subsection: nominal vehicle model} and a GP-based residual model in Section \ref{subsection: residual model}. Section \ref{subsection: analysis of nominal model discrepancies} presents the analysis of the nominal model error, which serves as the foundation for the residual model. More details can be found in \cite{li2025learning}.

\begin{table}[!t]
\normalsize
\caption{\textbf{Vehicle Model Notation}}
\label{table: vehicle model notation}
\centering
\begin{tabular}{cl}
\toprule
Symbols & Definition \\
\midrule
$X,Y$& Global position in cartesian coordinates.\\
$\varphi$& Yaw angle.\\
$v_x,v_y$& Longitudinal and lateral velocities of vehicle.\\
$r$& Yaw rate.\\
$T,\Delta T$& Command torque and its change rate.\\
$\delta,\Delta \delta$& Steering angle and its change rate.\\
$F_{f,x},F_{r,x}$& Longitudinal forces of front and rear tires.\\
$F_{f,y},F_{r,y}$& Lateral forces of front and rear tires.\\
$F_d$& Air drag.\\
$l_f,l_r$& Distance from C.G. to front and rear axles.\\
$m$& Vehicle mass.\\
$I_z$& Vehicle moment on inertia.\\
$\alpha_f,\alpha_r$& Front and rear wheel side slip angles.\\
$\kappa$& Torque distribution coefficient\\
$r_e$& Rolling radius of the tire.\\
$C_{f,r},C_{r,r}$& Rolling resistance of tire.\\
$C_w$& Coefficient of air drag.\\

\bottomrule
\end{tabular}
\end{table}


The considered control-oriented vehicle model in discrete time is formulated as: 
\begin{equation}
\label{equation: nominal and residual}
\bm{x}_{k+1} = \bm{f}(\bm{x}_k,\bm{u}_k)+\bm{B}_d\bm{g}(\bm{z}_k)
\end{equation}
where $\bm{x}=[X,Y,\varphi,v_x,v_y,r,T,\delta]^T$ is the states of vehicle and $\bm{u}=[\Delta T, \Delta \delta]^T$ is the control inputs, as defined in Table \ref{table: vehicle model notation}. $\bm{f}$ represents the nominal model and $\bm{g}$ is the residual model. 
The residual model is assumed to only affect the subspace spanned by $\bm{B}_d$. Feature vector $\bm{z}_k \in \mathbb{R}^{n_z}$ of residual model is extracted from $\bm{x}_k$ and $\bm{u}_k$.


\subsection{Nominal Vehicle Model}
\label{subsection: nominal vehicle model}

\begin{figure}[!t]
\centering
\includegraphics{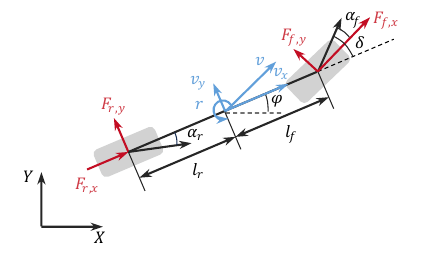}
\caption{The schematic of the single-track dynamics model. The velocities in vehicle states are depicted in blue, while the tire forces are represented in red.}
\label{figure: vehicle dynamics model}
\end{figure}

The nominal model employed in this work is selected as the single-track dynamics model, as illustrated in Fig. \ref{figure: vehicle dynamics model}. The system state equations of the single-track dynamics model are outlined as follows:
\begin{equation}
\label{equation: nominal model equation}
\bm{\dot{x}}
= \begin{bmatrix} 
v_x\mathrm{cos}\varphi-v_y\mathrm{sin}\varphi \\ 
v_x\mathrm{sin}\varphi+v_y\mathrm{cos}\varphi\\ 
r \\
\frac{1}{m}(F_{r,x}-F_{d}-F_{f,y}\mathrm{sin}\delta+F_{f,x}\mathrm{cos}\delta)+v_{y}r \\
\frac{1}{m}(F_{r,y}+F_{f,y}\mathrm{cos}\delta+F_{f,x}\mathrm{sin}\delta)-v_{x}r \\
\frac{1}{I_{z}}((F_{f,y}\mathrm{cos}\delta+F_{f,x}\mathrm{sin}\delta)l_f-F_{r,y}l_r) \\
\Delta T \\
\Delta \delta 
\end{bmatrix}
\end{equation}
All notations appearing in the equations are defined in Table \ref{table: vehicle model notation}. Lateral forces $F_{f,y}$ and $F_{r,y}$ are calculated using a simplified Magic Formula \cite{pacejka2005tire}:
\begin{equation}
\label{equation: magic formula}
\begin{aligned}
F_{f,y} &= D_{f}\mathrm{sin}(C_{f}\mathrm{arctan}(B_{f}\alpha_{f}))\\
F_{r,y} &= D_{r}\mathrm{sin}(C_{r}\mathrm{arctan}(B_{r}\alpha_{r}))
\end{aligned}
\end{equation}
The parameters $B_{f/r},C_{f/r},D_{f/r}$ are tire parameters used in the Magic Formula. The front and rear wheel side slip angles $\alpha_f$ and $\alpha_r$ are calculated based on kinematic relationships:
\begin{equation}
\label{equation: side slip angle}
\begin{aligned}
\alpha_f &= \mathrm{arctan}(\frac{v_y+l_fr}{v_x}) - \delta \\ 
\alpha_r &= \mathrm{arctan}(\frac{-v_y+l_rr}{v_x})
\end{aligned}
\end{equation}
The longitudinal tire forces $F_{f,x}$ and $F_{r,x}$ are modeled as: 
\begin{equation}
\label{equation: driving force}
\begin{aligned}
F_{f,x} &= \kappa \frac{T}{r_e}-C_{f,r} \\
F_{r,x} &= (1-\kappa)\frac{T}{r_e}-C_{r,r}
\end{aligned}
\end{equation}
The air drag $F_d$ is modeled as:
\begin{equation}
\label{equation: air drag}
F_{d} = C_w v_x^2
\end{equation}
When employed in the controller, the nominal model in (\ref{equation: nominal model equation}) is linearized and discretized, resulting in the nominal model in discret time $\bm{f}(\bm{x}_k,\bm{u}_k)$ as in (\ref{equation: nominal and residual}).

\subsection{Analysis of Nominal Model Discrepancies}
\label{subsection: analysis of nominal model discrepancies}
While the nominal model captures the essential vehicle dynamics, discrepancies remain compared to the actual dynamics.
Among the vehicle states, $X$, $Y$, and $\varphi$ are governed purely by kinematic relations, whereas \(T\) and \(\delta\) correspond to actuator outputs.
The dynamics of these states are assumed to be sufficiently accurate, with model errors mainly associated with velocity states $\begin{bmatrix}v_x,v_y,r\end{bmatrix}^\text{T}$.
Consequently, \(\bm{B}_d\) is defined as \(\bm{B}_d=\begin{bmatrix} \bm{0}_{3\times3} & \bm{I}_{3\times3} & \bm{0}_{3\times2} \end{bmatrix}^\text{T}\), and the dynamics of the velocity states can be expressed in matrix form as:
\begin{align}
\begin{bmatrix} \dot{v}_x \\ \dot{v}_y \\ \dot{r} \end{bmatrix} &= 
\begin{bmatrix} 
\cos\delta/m & -\sin\delta/m & 1/m & 0\\ 
\sin\delta/m & \cos\delta/m & 0 & 1/m \\ 
l_f\sin\delta/I_z & l_f\cos\delta/I_z & 0 & -l_r/I_z
\end{bmatrix}
\begin{bmatrix} F_{f,x} \\ F_{f,y} \\ F_{r,x} \\ F_{r,y} \end{bmatrix} \notag \\
&\quad +\begin{bmatrix} v_yr - C_wv_x^2 & -v_xr & 0 \end{bmatrix}^\text{T}
\nonumber
\label{dynamic state equation}
\end{align}

In the nominal vehicle model, the composition of forces and the parameters $m$, $l_f$, $l_r$, $I_z$, and $C_w$ are considered invariable elements, since they can be accurately calibrated and remain nearly constant during operation. 
Conversely, the tire forces $\begin{bmatrix}F_{f,x},F_{f,y},F_{r,x},F_{r,y}\end{bmatrix}^\text{T}$ are considered the variable elements, as they may deviate from nominal model during operation due to road conditions, temperature, and tire wear, while the coupling between longitudinal and lateral forces further contributes to deviations. These deviations constitute the primary source of modeling error.

Additionally, Considering that the front wheel steering angle $\delta$ is generally small, and the deviations in tire forces $\Delta F_{f/r,x/y}$ are also small relative to the values of tire model $F_{f/r,x/y}$, the impact of higher-order terms $\Delta F_{f/r,x/y} \sin\delta$ can be neglected, and $\cos\delta \approx 1$. Consequently, the expression for model error can be simplified as follows:
\begin{equation}
\label{final error dynamics equation}
\begin{bmatrix} \Delta\dot{v}_x \\ \Delta\dot{v}_y \\ \Delta\dot{r} \end{bmatrix} 
\approx 
\underbrace{\begin{bmatrix} 1/m & 0 & 1/m & 0\\ 0 & 1/m & 0 & 1/m \\ 0 & l_f/I_z & 0 & -l_r/I_z\end{bmatrix}}_{\text{invariable elements}}
\underbrace{\begin{bmatrix} \Delta F_{f,x} \\ \Delta F_{f,y} \\ \Delta F_{r,x} \\ \Delta F_{r,y} \end{bmatrix}}_{\text{variable elements}}
\end{equation}



\subsection{Residual Model}
\label{subsection: residual model}
To enable data-driven compensation for the nominal model, 
GPs are employed in the formulation of residual model, leveraging their flexibility and probabilistic nature to learn deviations caused by the variable elements of the nominal model.
Label $\bm{y}_k$ is derived by calculating the deviation between measurements and the prediction of the nominal model:
\begin{equation}
\label{deviations calculation}
\bm{y}_k = \bm{g}(\bm{z}_k)+\bm{w}_k = \bm{B}_d^{\dagger}(\bm{x}_{k+1}-\bm{f}(\bm{x}_k,\bm{u}_k))
\end{equation}
where $\bm{w}_k \sim \mathcal{N}(\bm{0},\bm{\Sigma}_w)$ is gaussian noise with variance $\bm{\Sigma}_w $= diag$(\begin{bmatrix}\sigma_{v_x}^2,\sigma_{v_y}^2,\sigma_{r}^2\end{bmatrix})$. $\bm{B}_d^{\dagger}$ is the Moore-Penrose pseudo-inverse of $\bm{B}_d$. 
The training set required for GPs is constructed from real-time measurements and labels:
\begin{equation}
\label{dataset}
\mathcal{D} = \begin{bmatrix}
    \bm{Z} = {\begin{bmatrix}
        \bm{z}_0^\text{T}; \cdots ;\bm{z}_{\lvert \mathcal{D} \rvert}^\text{T}
    \end{bmatrix}},
    \bm{Y} = {\begin{bmatrix}
        \bm{y}_0^\text{T}; \cdots ;\bm{y}_{\lvert \mathcal{D} \rvert}^\text{T}
    \end{bmatrix}}
\end{bmatrix}
\end{equation}

Furthermore, each dimension of the residual model output is assumed to be uncorrelated with the others. Under this assumption, GPs can provide posterior predictions of the mean $\bm{\mu}$ and variance $\Sigma$ of the nominal model error at test point $\bm{z}_*$:
\begin{equation}
\begin{aligned}
\label{equation: gp prediction}
    \bm{\mu}(\bm{z}_*) &= \bm{k}_*^\text{T} (\bm{K} + \sigma^2 \bm{I})^{-1} \bm{Y} \\
    \Sigma(\bm{z}_*) &= k(\bm{z}_*, \bm{z}_*) - \bm{k}_*^\text{T} (\bm{K} + \sigma^2 \bm{I})^{-1} \bm{k}_*
\end{aligned}
\end{equation}
The matrix $\bm{K}$ denotes the covariance evaluated at all pairs of points in training set and $\begin{bmatrix} \bm{K} \end{bmatrix}_{ij} = k(\bm{z}_i,\bm{z}_j)$. The vector $\bm{k}_* = \begin{bmatrix} k(\bm{z}_*,\bm{z}_1), \cdots ,k(\bm{z}_*,\bm{z}_m) \end{bmatrix}^\text{T}$. The Squared Exponential function is employed to define the GP kernel $k(\bm{z}_i,\bm{z}_j)$:
\begin{equation}
\label{SE kernel function}
k(\bm{z}_i,\bm{z}_j) = \sigma_f^2 \text{exp}(-\frac{1}{2}(\bm{z}_i-\bm{z}_j)^\text{T}\bm{M}(\bm{z}_i-\bm{z}_j))
\end{equation}
where parameter $\bm{M}$ defines the length-scale matrix and parameter $\sigma_f^2$ defines the squared signal variance.
The final multivariate GPs approximation of the unknown residual $\bm{g}(\bm{z}_*)$ is obtained by aggregating the outputs for each dimension:
 \begin{equation}
     \bm{g}(\bm{z}_*) \sim \mathcal{N}(\bm{\mu},\bm{\Sigma})
 \end{equation}
 where $\bm{\mu} = \begin{bmatrix} \mu_{v_x},\mu_{v_y},\mu_r \end{bmatrix}^\text{T}$ and $\bm{\Sigma} = $diag$(\begin{bmatrix}\Sigma_{v_x},\Sigma_{v_y},\Sigma_r\end{bmatrix}^\text{T})$.

As indicated in (\ref{final error dynamics equation}), the dominant source of error originates from discrepancies between the actual tire forces and those predicted by the tire model (\ref{equation: magic formula}).
In the nominal model, lateral tire forces are functions of tire slip angles, implying that the discrepancies $\Delta F_{f/r,y}$ are primarily related to $\alpha_{f/r}$. 
Similarly, longitudinal tire force deviations $\Delta F_{f/r,x}$ are primarily caused by the mismatch between the commanded torque and the actual longitudinal force responses, rendering $\Delta F_{f/r,x}$ largely dependent on $T$. 
Furthermore, accounting for the coupling between longitudinal and lateral tire forces, the overall deviations $\Delta F_{f/r,x/y}$ depend jointly on both the slip angles and the commanded torque. 
Consequently, the front and rear slip angles $\alpha_f$, $\alpha_r$, together with the commanded torque $T$, are selected as the input features for the residual model:
\begin{equation}
\label{residual model input features}
\bm{z} = \begin{bmatrix} \alpha_f, \alpha_r, T\end{bmatrix}^\text{T}
\end{equation}

In existing researches, residual models are typically designed to capture all potential sources of modeling error, and thus employ the dynamic states $\begin{bmatrix} v_x, v_y, r, \delta, T \end{bmatrix}^\text{T}$ as input features. These features span a 5-dimensional feature space, which necessitates a vast training set for sufficient coverage.
Additionally, it is infeasible to control the vehicle to traverse every region within this 5-dimensional space.
In contrast, the residual model in this article only compensate for the error arising from the variable elements.
Under this targeted formulation and reasonable assumptions, the input features are reduced to 3 variables. 
Although tire slip angles are not included in the vehicle states, they can be directly derived via simple equations in (\ref{equation: side slip angle}).

This dimensionality reduction from 5 to 3 significantly decreases the required training set size, enabling sufficient coverage of the entire vehicle performance envelope using a tractable training set, as further elaborated in the next section.

\section{Sparse Incremental Learning of Error Dynamics}
\label{section: sparse incremental learning of error dynamics}

\begin{figure*}[!t]
\centering
\includegraphics{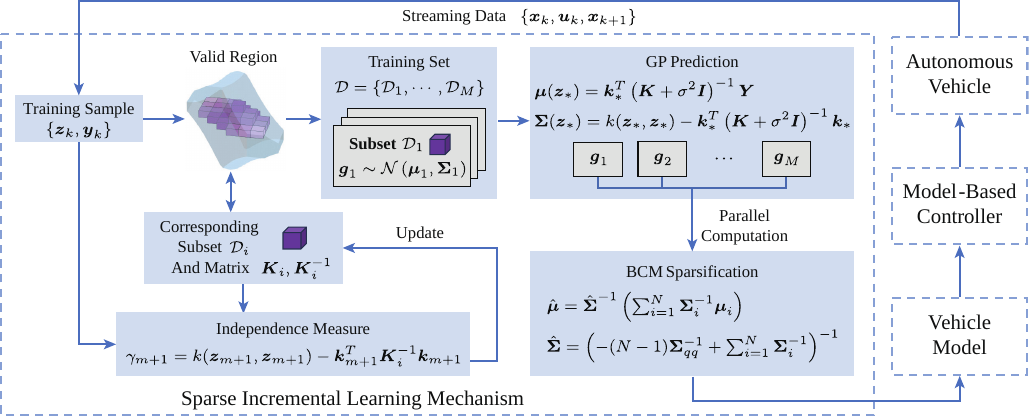}
\caption{Framework of sparse incremental learning. At each sampling cycle, the streaming data are used to update the data subsets within the corresponding subregion in the valid region. The GP predictions from all subsets are aggregated via BCM to estimate the model errors. The estimation is then incorporated into the vehicle model and used by the model-based controller to compute the control inputs, which are subsequently applied to the autonomous vehicle.}
\label{figure: SPLIT framework}
\end{figure*}

In this section, we introduce the sparse incremental learning mechanism, as illustrated in Fig. \ref{figure: SPLIT framework}. Section \ref{subsection: valid region definition} explicitly defines the valid region within the feature space that the training set is expected to cover. 
Section \ref{subsection: sparse incremental learning} then introduces the sparse incremental learning mechanism within the valid region.
Section \ref{subsection: sparse gaussian process} introduces the sparse approximation methods BCM for GPs.


\subsection{Valid Region Definition}
\label{subsection: valid region definition}

\begin{figure}[!t]
\centering
\includegraphics{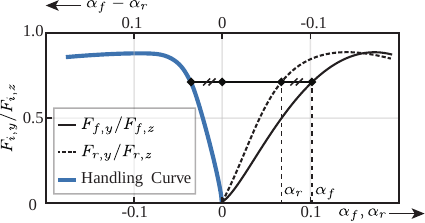}
\caption{Illustration of the handling curve. The right side shows the effective axle characteristics \(F_{f,y}(\alpha_f)/ F_{f,z}\) and \(F_{r,y}(\alpha_r)/ F_{r,z}\) of the front and rear axles, while the left side presents the handling curve derived from them. The handling curve is a function of \(\alpha_f-\alpha_r\), which increases with the slip angle difference up to a threshold and then ceases to grow.}
\label{figure: handling diagram}
\end{figure}

\begin{figure*}[!t]
\centering
\includegraphics{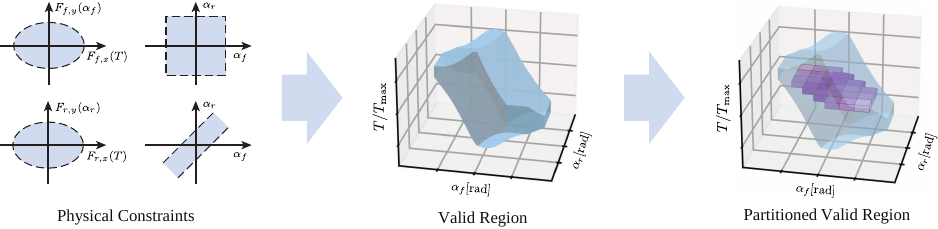}
\caption{Illustration of the definition and partitioning of the valid region. The valid region (blue) is explicitly defined as the intersection of three sets of physical  constraints among the residual model features, delineating the attainable region of the feature space during vehicle operation. It is then uniformly partitioned into equal-sized cubic subregions (purple), with each subregion maintaining a local data subset that collectively forms the training set of the residual model. During online updates, streaming data are used only to update the subset corresponding to their subregion, making the update process computationally tractable in real time. This partitioning and localized updating guarantee uniform and sufficient coverage of the valid region by the training set.}
\label{figure: valid region definition}
\end{figure*}

Although the dimensionality of the feature space is reduced, it remains both infeasible and unnecessary to cover the entire feature space. 
This is due to the fact that features have concrete physical meanings and are naturally bounded. They are subject to strict physical constraints that limit their admissible ranges, rendering most regions within the feature space physically infeasible.
As a result, the region of the feature space that is physically attainable by the vehicle is confined to a small neighborhood near the origin, referred to as the valid region, which the training set of residual model is expected to adequately cover.
Here, three sets of inequality constraints among selected features are introduced, which are employed to explicitly define the valid region within the feature space. 
Data violating these constraints are considered invalid and excluded.

The first constraint concerns tire force limits. 
To ensure safety, the combined longitudinal and lateral forces of the front and rear tires should be constrained within the tire-road contact ellipse to prevent instability and hazardous conditions:
\begin{equation}
\label{equation: tire forces constraint}
\begin{aligned}
    (p_{\text{long}}F_{f,x})^2+F_{f,y}^2 \leq (p_{\text{ellipse}}D_{f})^2\\
    (p_{\text{long}}F_{r,x})^2+F_{r,y}^2 \leq (p_{\text{ellipse}}D_{r})^2
\end{aligned}
\end{equation}
with $p_{\text{long}}$ and $p_{\text{ellipse}}$ defining the ellipse shape.

The second constraint concerns the tire slip angle limits. By restricting the maximum slip angles, this constraint seeks to prevent tires from reaching the lateral force saturation zone:
\begin{equation}
\label{equation: slip angle constraint}
\begin{aligned}
    \lvert \alpha_{f} \rvert \leq \alpha_{\text{max}}\\
    \lvert \alpha_{r} \rvert \leq \alpha_{\text{max}}
\end{aligned}
\end{equation}

The third constraint is derived from the vehicle handling diagram \cite{rossa2012bifurcation}, as illustrated in Fig. \ref{figure: handling diagram}, which characterizes the relationship between the effective axle characteristics \(F_{f/r,y}/ F_{f/r,z}\) and the corresponding tire slip angles. As evident from these curves, beyond a certain threshold, further increases in the difference \(\alpha_f - \alpha_r\) cease to enhance the lateral acceleration. Consequently, \(\alpha_f - \alpha_r\) should be appropriately constrained within a physically reasonable range:
\begin{equation}
\label{equation: difference of slip angle constraint}
\lvert \alpha_f - \alpha_r \rvert \leq \Delta \alpha_{\text{max}}
\end{equation}

Building on three sets of constraints, the valid region which encompasses the majority of the dynamic states encountered during operation within the feature space is explicitly defined, as depicted in the Fig. \ref{figure: valid region definition}.

\subsection{Sparse Incremental Learning}
\label{subsection: sparse incremental learning}

\begin{figure}[!t]
\centering
\includegraphics{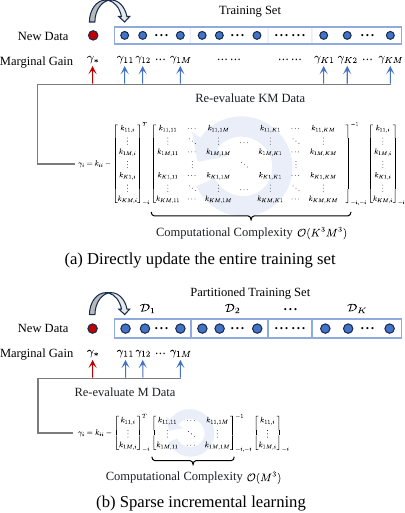}
\caption{Comparison between directly updating the entire training set and sparse incremental learning. Red and blue circles represent new and existing data. For a training set with $KM$ samples partitioned into $K$ subsets of size $M$, direct updates require $\mathcal{O}(K^3M^3)$ complexity and $KM$ re-evaluations of marginal gains, since each new sample affects all existing ones. In contrast, sparse incremental learning updates only the subset within the corresponding subregion, reducing the complexity to $\mathcal{O}(M^3)$ with only $M$ re-evaluations. Thus, SPLIT reduces both the computational cost of each evaluation and the total number of evaluations, making real-time online learning feasible.}
\label{figure: sparse incremental learning}
\end{figure}

By defining the valid region, the challenge of online learning the error dynamics across the entire performance envelope is transformed into ensuring sufficiently coverage of valid region, which requires updating the training set with data that provide the highest marginal gain in feature space coverage. The inclusion of new samples and the removal of redundant ones alter the marginal gains of all other data. As a result, each update necessitates a full re-evaluation of the marginal gain for every sample in the training set, leading to considerable redundant computation.
We observe that the marginal gain of a given data to coverage is primarily governed by its spatial proximity to other data.
Specifically, each sample exerts significant influence on the marginal gains of neighboring data, while its effect on distant samples remains negligible.
This observation motivates the restriction of training set updates to localized subregions to enable the learning process online.

Therefore, the valid region is partitioned into \( K \) equal sized cubic subregions, within which a data subset $\mathcal{D}_k$ of size $M$ is established for independent data updates. 
The partitioned valid region within feature space is illustrated in Fig. \ref{figure: valid region definition}. When new measurements are acquired in real-time, the initial step involves identifying the specific cubic subregion that the new data occupies. Following this, the corresponding data subset is updated with the new measurements. 

The marginal gain in feature space coverage of a new data $\bm{d}_{m+1}=(\bm{z}_{m+1},\bm{y}_{m+1})$ is quantified by the independence measure, as delineated in \cite{nguyen2011incremental}, which is formally defined as follows:
\begin{equation}
\label{equation: independence measurement}
\gamma_{m+1} = k(\bm{z}_{m+1},\bm{z}_{m+1}) - \bm{k}_{k,m+1}^T\bm{K}_k^{-1}\bm{k}_{k,m+1}
\end{equation}
Here, matrix $\bm{K}_k$ denotes the kernel matrix in subset $\mathcal{D}_k$, where \( [\bm{K}_k]_{ij} = k(\bm{z}_i, \bm{z}_j) \), and vector $\bm{k}_{k,m+1}$ represents the kernel vector, with \( [\bm{k}_{k,m+1}]_i = k(\bm{z}_i, \bm{z}_{m+1}) \). A larger value of $\gamma_{m+1}$ indicates a higher marginal gain of the new data, implying that it provides additional valuable information for model fitting. 
The marginal gain \(\gamma_i\) is maintained for each data \(\bm{d}_i\) within subset \(\mathcal{D}_k\), which is evaluated with respect to the remaining data \(\mathcal{D}_k \setminus \{\bm{d}_i\}\) in the subset. Specifically, the kernel matrix \(\bm{K}_k\) and kernel vector \(\bm{k}_k\) in (\ref{equation: independence measurement}) are replaced by \(\bm{K}_k^i\) and \(\bm{k}_k^i\), which are computed based on \(\mathcal{D}_k \setminus \{\bm{d}_i\}\) and \(\bm{d}_i\).

If the size of \(\mathcal{D}_k\) has not yet reached the predefined maximum \(M\), and the marginal gain of the new data exceeds the specified threshold, the subset \(\mathcal{D}_k\) is directly augmented by including the new data \(\mathcal{D}_k = \mathcal{D}_k \cup \{\bm{d}_{m+1}\}\).
As a preliminary step, the re-evaluation of the marginal gain for existing samples necessitates updating the intermediate variables \(\bm{K}_k^i\) and \(\bm{k}_k^i\) associated with each data \(\bm{d}_i\) by augmenting them with additional dimensions: 
\begin{equation}
\label{equation: K_i update}
\begin{aligned}
\bm{K}_{k,\text{new}}^i = \begin{bmatrix}  \bm{K}_{k,\text{old}}^i & \bm{k}_{k,m+1}^i \\ \bm{k}_{k,m+1}^{iT} & k_{m+1}\end{bmatrix}
\\\bm{k}_{k,\text{new}}^i=\begin{bmatrix}  \bm{k}_{k,\text{old}}^i & k_{i,m+1} \end{bmatrix}^T
\end{aligned}
\end{equation}
\allowdisplaybreaks
where $\bm{k}_{k,m+1}^i=[\bm{k}_{k,m+1}]_{\setminus i}$, $k_{m+1}=k(\bm{z}_{m+1},\bm{z}_{m+1})$, and $k_{i,m+1}=k(\bm{z}_{i},\bm{z}_{m+1})$. The notation $[\bm{k}_{k,m+1}]_{\setminus i}$ denotes the vector obtained by removing the $i$-th element from the vector $\bm{k}_{k,m+1}$. 
Subsequently, the updated marginal gain $\gamma_i$ is calculated by inserting the updated intermediate variables $\bm{K}_{k,\text{new}}^i$ and $\bm{k}_{k,\text{new}}^i$ into the formulation defined in (\ref{equation: independence measurement}).
The kernel $\bm{K}_k$ in subset $\mathcal{D}_k$ is also updated in a manner analogous to that in (\ref{equation: K_i update}).

When $\mathcal{D}_k$ reaches its maximum capacity $M$, and the marginal gain of the incoming data exceeds the minimum marginal gain among existing samples in $\mathcal{D}_k$, the data with the lowest marginal gain (indexed as $l$) will be removed and replaced by the new data.
The intermediate variables $\bm{K}_k^i$ and $\bm{k}_k^i$ of remaining points 
are updated by replacing the $l$-th row and column of the matrix to re-evaluate the $\gamma_i$ as in (\ref{equation: independence measurement}).
Although the inverse of $\bm{K}_k^i$ can be computed incrementally using Cholesky decomposition \cite{nguyen2011incremental}, repeated updates may accumulate numerical errors and compromise computational accuracy.
Since the data updates are performed exclusively within the subset of size \(M\), the matrix $\bm{K}_k^i$ is directly inverted to maintain precision and the computational burden is considerably diminished.

The ability of SPLIT to perform online learning and maintain a large-scale training set in real time is primarily attributed to two principal factors, as illustrated in fig.\ref{figure: sparse incremental learning}. 

\begin{enumerate}
\item Computational complexity of each marginal gain evaluation is reduced by decreasing the dimension of kernel \(\bm{K}_k^i\) from approximately \(KM\) to \(M\), which reduces the cost of matrix inversion from $\mathcal{O}(K^3M^3)$ to $\mathcal{O}(M^3)$.
\item SPLIT limits the number of marginal gain evaluations to \(M\) samples within the subregion, rather than almost \(KM\) samples in the full training set. This further reduces the total number of evaluations by a factor of \(K\).

\end{enumerate}


\subsection{Sparse Gaussian Processes}
\label{subsection: sparse gaussian process}

\begin{figure}[!t]
\centering
\includegraphics{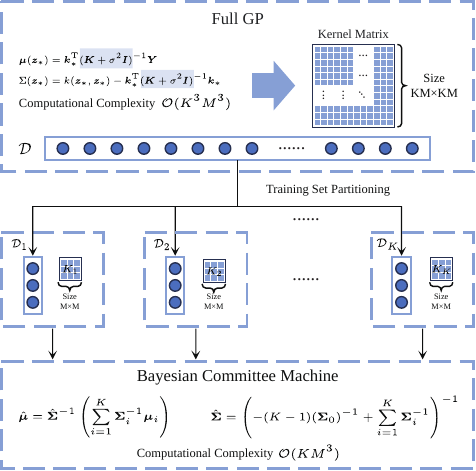}
\caption{Illustration of GP sparsification using BCM. For a training set with $KM$ samples partitioned into $K$ subsets of size $M$, a full GP requires inverting a large $KM \times KM$ kernel matrix with complexity $\mathcal{O}(K^3M^3)$. In contrast, BCM trains a separate GP on each subset and integrates the predictions to form the final result, requiring inversion of $K$ small $M \times M$ kernel matrices, which reduces the complexity to $\mathcal{O}(KM^3)$ and enables efficient online evaluation.}
\label{figure: bcm}
\end{figure}

While the sparse incremental learning mechanism ensures sufficient coverage of the valid region, it simultaneously results in an excessively large training set, leading to the intractable computational complexity of the online evaluation. To meet the real-time requirements of the vehicle controller, we use the BCM to simplify the online evaluation of GP. 
The core principle of BCM is partitioning the entire training set into multiple subsets, with each subset serving as the training set for a distinct GP. 
This partitioning of the training set aligns seamlessly with the partitioning strategy employed for the valid region as outlined in Section \ref{subsection: sparse incremental learning}. 

Accordingly, the $K$ subsets \( \mathcal{D} = \{\mathcal{D}_1, \cdots, \mathcal{D}_K\} \) derived from the training set partitioning are leveraged for computations within the framework of the BCM, each of which is employed to train an individual GP $\bm{g}_i \sim \mathcal{N}(\bm{\mu}_i, \bm{\Sigma}_i), i = 1, \cdots, K $.
Let $D^i = \{\mathcal{D}_1, \cdots, \mathcal{D}_i\}$ represent the set of all subsets with indices less than or equal to $i$, where $i = 1, \cdots, K$. 
Let $P(\bm{g} | \mathcal{D}_i)$ represent the posterior probability density function derived from the GP corresponding to the subset $\mathcal{D}_i$ at test points. The following expression holds:

\begin{equation}
\label{equation: posterior predictive probability density}
\begin{aligned}
    P(\bm{g} | \mathcal{D}) &= P(\bm{g} | D^{i-1}, \mathcal{D}_i) \\ &\propto P(\bm{g}) P(D^{i-1} | \bm{g}) P(\mathcal{D}_i | D^{i-1}, \bm{g})
\end{aligned}
\end{equation}
The key to BCM sparse approximation of GPs is the introduction of a critical assumption:
\begin{equation}
\label{equation: key assumption}
P(\mathcal{D}_i | D^{i-1}, \bm{g}) \approx P(\mathcal{D}_i | \bm{g})
\end{equation}
In general, this assumption does not hold. It is only reasonable when the correlation between $D^{i-1}$ and $\mathcal{D}_i$ is weak, meaning the association or influence between individual subsets and the remaining subsets is minimal.
In our formulation, the subsets derived from partitioning the valid region are spatially disjoint, resulting in negligible inter-subset correlation. This structural separation ensures that the independence assumption underpinning BCM is approximately satisfied.

Utilizing this assumption and applying Bayes' theorem to $P(D^{i-1} | \bm{g})$ and $P(\mathcal{D}_i | \bm{g})$, the posterior probability $P(\bm{g} | D^{i-1}, \mathcal{D}_i)$ can be approximated as:
\begin{equation}
\label{equation: Bayes' theorem}
P(\bm{g} | D^{i-1}, \mathcal{D}_i) \propto \frac{P(\bm{g} | D^{i-1}) P(\bm{g} | \mathcal{D}_i)}{P(\bm{g})}
\end{equation}
And the approximate posterior probability distribution of the residual model is derived as:
\begin{equation}
\label{equation: approximate posterior probability distribution}
P(\bm{g} | \mathcal{D}) \propto \frac{\prod_{i=1}^K P(\bm{g} | \mathcal{D}_i)}{P(\bm{g})^{K-1}}
\end{equation}
Assuming a prior probability $P(\bm{g}) = \mathcal{N}(0, \bm{\Sigma}_{0})$, the result after sparse approximation is a Gaussian distribution with mean $\hat{\bm{\mu}}$ and covariance matrix $\hat{\bm{\Sigma}}$:

\begin{equation}
\label{equation: bcm prediction}
\begin{aligned}
&\hat{\bm{\mu}} = \hat{\bm{\Sigma}}^{-1} \left( \sum_{i=1}^K \bm{\Sigma}_i^{-1} \bm{\mu}_i \right) \\
&\hat{\bm{\Sigma}} = \left( - (K-1)(\bm{\Sigma}_{0})^{-1} + \sum_{i=1}^K \bm{\Sigma}_i^{-1} \right)^{-1}
\end{aligned}
\end{equation}
The ultimate sparse prediction of BCM is derived by computing the weighted average of the GP predictions from each GP, with weights determined by the inverse of their respective covariance matrices. Predictions from subsets located near the test point exhibit reduced variance, thus exerting a greater influence on the final prediction.


As shown in fig \ref{figure: bcm}, the application of BCM reduces the computational complexity of online GP evaluation from \(\mathcal{O}(K^3M^3)\) to \(\mathcal{O}(KM^3)\), where \(KM\) denotes the total training set size, achieving linear scalability with respect to training set size and enabling efficient large-scale online inference.
Since GP models within each subregion operate independently, BCM naturally supports parallel execution across multiple threads, further improving computational efficiency.
Moreover, the kernel matrix \(\bm{K}_i\) and its inverse \(\bm{K}_i^{-1}\) for each subset \(\mathcal{D}_i\) are precomputed during sparse incremental learning and reused in the BCM prediction (\ref{equation: bcm prediction}), which eliminates repeated matrix inversions.
Given that the size of each subset \( M \) is relatively small, the computational burden of the sparse representation is substantially lower than that of full GPs.

\section{Model-Based Controller Formulation}
\label{section: model-based controller formulation}

In this section, two model-based controllers are designed and integrated with SPLIT to assess its effectiveness in enhancing controller performance.

\subsection{Model Predictive Control}
\label{subsection: model predictive control}

The first controller is a standard MPC, employed to evaluate the performance improvement of SPLIT in the context of standard trajectory tracking. The optimization objective is to minimize a quadratic cost penalizing the deviations of the predicted states $\bm{x}_k$ and control inputs $\bm{u}_k$ from their corresponding reference values $\bm{x}_{k,\text{ref}}$ and $\bm{u}_{k,\text{ref}}$. The formulation of the controller is as follows:

\begin{equation}
\begin{aligned}
\min_{\{\bm{u}_k\}}  \;& \sum_{k=1}^H (\Vert \Delta \bm{x}_k \Vert_{\bm{Q}}^2 + \Vert \Delta \bm{u}_k \Vert_{\bm{R}}^2) + \Vert \Delta \bm{x}_H \Vert_{\bm{P}}^2\\
    \text{s.t.} \
   & \bm{x}_0 = \bm{x} \\
   & \bm{x}_{k+1} = \bm{f}(\bm{x}_k, \bm{u}_k) + \bm{B}_d\bm{g}(\bm{z}_k) \\
   & \bm{x}_k \in \mathcal{X}, \; \bm{u}_k \in \mathcal{U}
\end{aligned}
\end{equation}

Here, $\Delta \bm{x}_k = \bm{x}_k - \bm{x}_{k,\text{ref}}$ and $\Delta \bm{u}_k = \bm{u}_k - \bm{u}_{k,\text{ref}}$, where $\bm{Q}\geq0$, $\bm{R}>0$, and $\bm{P}\geq0$ denote the weighting matrices for the state, the input, and the terminal state, respectively. The notation $\|\cdot\|_{\bm{A}}^2$ represents the weighted Euclidean inner product, defined as $\|\bm{v}\|^2_{\bm{A}} = \bm{v}^T \bm{A} \bm{v}$.

\subsection{Model Predictive Contouring Control}
\label{subsection: model predictive contouring control}

The second controller is the Model Predictive Contouring Control (MPCC) controller. The state variables in (\ref{equation: nominal model equation}) are augmented with two additional states, $\theta$ and $v_s$, which approximate the progression of vehicle along the track and velocity along the centerline. The deviations between the vehicle position and the reference point on the track centerline are decomposed into the lag error $e_l$ and contour error $e_c$, defined as follows:
\begin{equation}
\label{equation: lag and contour error}
\begin{aligned}
e_l(\bm{x}_k,\theta_k)= -&\mathrm{cos}(\Phi(\theta_k))(X_k-X_c(\theta_k)) \\
-&\mathrm{sin}(\Phi(\theta_k))(Y_k-Y_c(\theta_k)) \\
e_c(\bm{x}_k,\theta_k)= \quad&\mathrm{sin}(\Phi(\theta_k))(X_k-X_c(\theta_k)) \\
-&\mathrm{cos}(\Phi(\theta_k))(Y_k-Y_c(\theta_k))
\end{aligned}
\end{equation}

MPCC integrates trajectory planning and tracking into a unified optimization problem and maximizes the progress along the track while respecting feasible track boundaries. The formulation of MPCC is as follows:
\begin{equation}
\label{equation: mpcc formulation}
\begin{aligned}
\min_{\{\bm{u}_k,v_k\}} &\sum_{k=1}^Hq_l \hat{e}_l(\hat{\theta}_k)^2 + q_c \hat{e}_c(\hat{\theta}_k)^2-q_v v_k+\Vert\bm{u}_k\Vert_{\bm{R}_{\bm{u}}}^2\\
 \text{s.t.} \;
    & \bm{x}_0 = \bm{x} \\
    & \bm{x}_{k+1} = \bm{f}(\bm{x}_k, \bm{u}_k) + \bm{B}_d\bm{g}(\bm{z}_k) \\
    &\theta_{k+1}=\theta_k+v_k \Delta t\\
    &\|X_k - X_c(\hat{\theta}_k)\|^2 + \|Y_k - Y_c(\hat{\theta}_k)\|^2 \leq R^2 \\
    &(p_{\text{long}}F_{f/r,x})^2+F_{f/r,y}^2 \leq (p_{\text{ellipse}}D_{f/r})^2\\
    &-\alpha_{\text{max}} \leq \alpha_{f/r} \leq \alpha_{\text{max}}\\
    &-\Delta \alpha_{\text{max}} \leq \alpha_f - \alpha_r \leq \Delta \alpha_{\text{max}}
\end{aligned}
\end{equation}

The term $q_v v_k$ is designed to maximize the vehicle progression along the track within the prediction horizon. Here, $q_l$, $q_c$, and $q_v$ are constant weights, and $\bm{R}_{\bm{u}}$ is the weighting matrix.
The trajectory constraint confines the vehicle position within the defined track radius \( R \). Furthermore, The three sets of constraints defined in (\ref{equation: tire forces constraint}), (\ref{equation: slip angle constraint}) and (\ref{equation: difference of slip angle constraint}) restrict the vehicle states within the valid region, mitigating the risk of hazardous behaviors. To guarantee the feasibility of the optimization problem, all constraints are modeled as soft constraints.

\section{Simulation Experiments}
\label{section: simulation experiments}

In this section, we present quantitative analysis based on the simulation. First, we evaluate the efficacy of SPLIT on overall control performance enhancement through an aggressive autonomous racing task. Then, we conduct ablative studies to quantify the contribution of the hybrid vehicle model, sparse incremental learning, and GP sparsification.




\subsection{Experiment Setup}
\label{subsection: experiment setup}

The experiment is conducted by racing the vehicle around a racetrack. The Class-B model within Carsim is employed as the simulation vehicle and the Handling Course race line, illustrated in Fig. \ref{figure: control performance enhancement}, serves as the racetrack, encompassing multiple curves and straight segments over a total distance of 2327.98 m.
The racing experiment consisted of 10 laps in total, with the data accumulated in the training set at the end of each lap being directly utilized as the initial training set for the subsequent lap. To mitigate the risk of irregular data, the learning mechanism are performed exclusively when the vehicle speed exceeds 5 m/s. Meanwhile, we also conduct a full lap of racing using the physical nominal controller as a comparison.
During the experiment, the maximum torque is set to 1000 N·m, the maximum vehicle speed and the initial vehicle speed are set to 30 m/s.


\subsection{Controller Implementation}
\label{subsection: controller implementation}
The controller employed in the experiment is the MPCC controller in (\ref{equation: mpcc formulation}), which operates with a sampling time of $\Delta t = 50 \,\text{ms}$ and a prediction horizon length of $H = 80$.
The associated optimization problem is solved using the HPIPM solver \cite{frison2020hpipm}, with the maximum iteration count set to 40, ensuring that the controller can be solved in real-time.
The constraint bounds are defined with the maximum tire slip angle $\alpha_{\text{max}} = 0.18 \,\text{rad}$, and the maximum difference between front and rear tire slip angles $\Delta \alpha_{\text{max}} = 0.10 \,\text{rad}$. 
The valid region within feature space is partitioned 
using cubic cells with edge lengths of $0.02 \,\text{rad}$, $0.02 \,\text{rad}$, and 0.1 along each feature dimension, respectively. In each subregion, a data subset \( \mathcal{D}_i \) containing \( M = 10 \) data points is maintained.
The controller is implemented in C++ and deployed on a Nuvo-9160 industrial computing platform, equipped with an Intel i9-13900 CPU. Control signals and state information of vehicle in simulation environment are transmitted via TCP communication protocols.

\subsection{Quantitative Evaluation of Control Performance Enhancement}
\label{subsection: quantitative evaluation of control performance enhancement}

\begin{figure*}[!t]
    \centering
\includegraphics{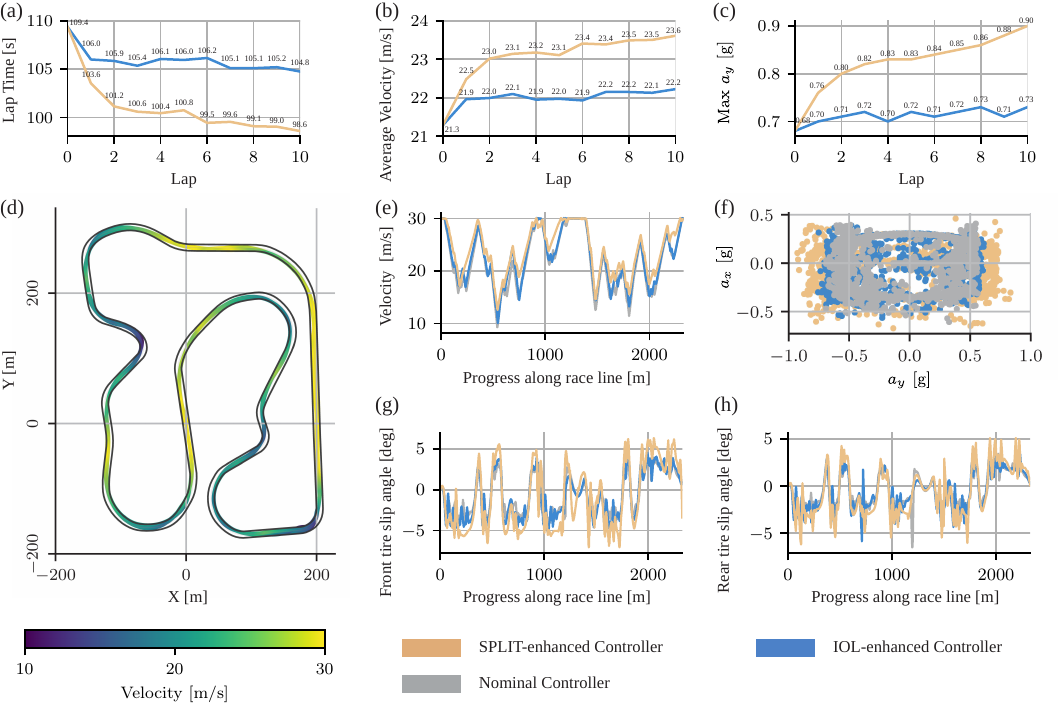}
\caption{Comparison of closed-loop simulation results with SPLIT-enhanced controller and IOL-enhanced controller during the entire autonomous racing task. (a)–(c) Comparison of lap time, average velocity, and performance improvement across laps for both learning methods. (d) Race track layout and closed-loop trajectory obtained by the SPLIT-enhanced controller during the entire racing. (e)–(h) Detailed comparison of the best laps achieved by SPLIT-enhanced and IOL-enhanced controllers, together with the baseline lap from the nominal controller, in terms of velocity profile, GG-diagram, front tire slip angle, and rear tire slip angle. Both error dynamics learning mechanism outperform the nominal controller, but IOL-enhanced controller exhibits diminishing learning effects after lap 3, whereas SPLIT demonstrates sustained improvements and superior handling of vehicle dynamics.}
\label{figure: control performance enhancement}
\setlength{\abovecaptionskip}{0.cm}
\end{figure*}

To evaluate the performance improvement brought by SPLIT, we compare it against the representative GP-based error dynamics online learning method in \cite{kabzan2019learning}, which has been shown to enhance control performance in various autonomous driving control tasks \cite{jiang2023fast, zhao2024learning}.
Notably, due to its high computational complexity, the baseline method cannot be executed in real-time and is therefore implemented in a separate thread, referred to as irregular online learning (IOL).
The FITC-based sparsification technique \cite{kabzan2019learning} is excluded from the online evaluation, as it introduces oscillations that lead to failure of the racing task.

Fig. \ref{figure: control performance enhancement} presents the experimental results during the entire racing.
Lap 0 serves as the baseline achieved through the nominal controller, while subsequent laps reflect results from the controller with the online learning mechanism.
Both online learning approaches demonstrate the ability to improve control performance over time. Compared to the nominal controller, they achieve a progressive increase in average velocity and the maximum lateral acceleration, alongside a reduction in lap time, collectively indicating enhanced capability in handling complex and nonlinear vehicle dynamics.
Notably, despite identical initial conditions and control architecture, SPLIT exhibits more efficient adaptation to the error dynamics, leading to consistently superior performance compared to the IOL baseline.
In lap 1, SPLIT reduces the lap time from 109.35 s achieved by the nominal controller to 103.55 s, already outperforming the best lap time 104.75 s achieved by the IOL baseline in lap 10. 
Over subsequent laps, the control performance continues to improve, attaining a minimum lap time of 98.6 s in lap 10, representing a 10.11\% improvement over the nominal controller.
The maximum lateral acceleration of 0.9 g in lap 10 indicates that the vehicle operates close to the tire–road friction limit, demonstrating that SPLIT fully exploits vehicle dynamics. In contrast, the learning efficacy of the IOL declines after lap 3, with lap times stabilizing around 105 s and the maximum lateral acceleration remaining notably lower than that attained with SPLIT.

Fig. \ref{figure: control performance enhancement} also compares the best lap data achieved by SPLIT, IOL, and lap 0 from nominal controller.
The results demonstrate SPLIT achieves the most significant enhancement in velocity at nearly every race track segment.
The GG-diagram for SPLIT exhibits a wider distribution in the lateral acceleration direction, indicating more effective utilization of tire-road friction.
These improvements confirm the superior capability of SPLIT to capture vehicle dynamics more accurately and enable the controller to exploit tire–road adhesion more effectively, which is corroborated by the increased ranges of front and rear tire sideslip angles.

The enhanced controller performance achieved by SPLIT arises from the synergistic effect of its constituent components.
Therefore, we evaluate the efficacy of each module in isolation in the following.

\subsection{Compensation for Model Error in Variable Elements}
\label{subsection: compensation for model error in variable elements}

\begin{figure*}[!t]
    \centering
\includegraphics{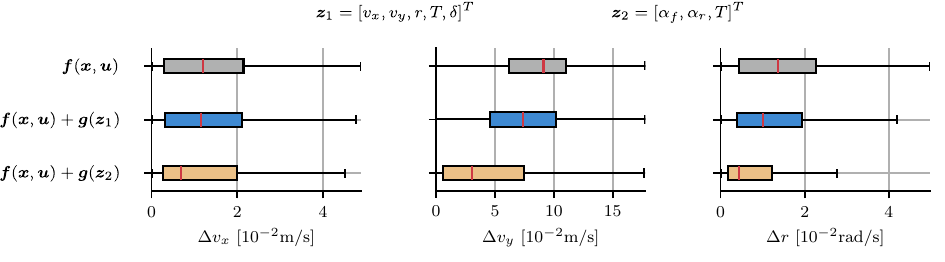}
\vspace{-5pt}
\caption{Comparison of model prediction errors with different feature selections for the residual model during lap 0. The training sets are respectively constructed with 100 most informative samples selected from lap 0 data using the marginal gain criterion defined in (\ref{equation: independence measurement}). With the same training set size, the low-dimensional features adopted in SPLIT provide broader coverage of operating conditions, leading to overall improvement in model accuracy.}
\label{figure: compensation for model error in variable elements}
\setlength{\abovecaptionskip}{0.cm}
\end{figure*}

\begin{figure*}[!t]
    \centering
\includegraphics[]{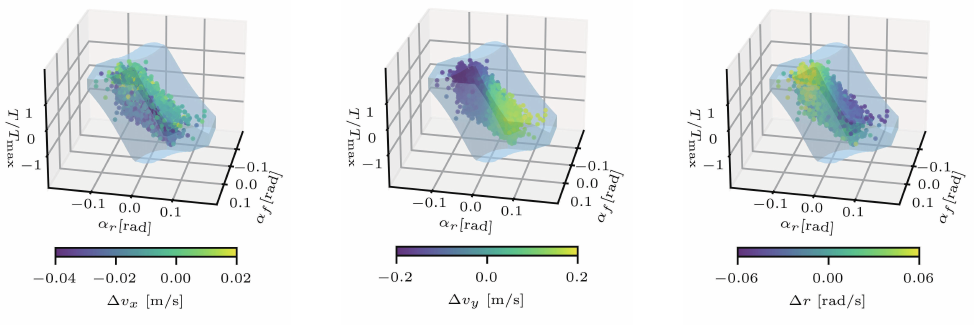}
\vspace{-5pt}
\caption{Distribution of physical model errors in velocity states within the feature space, based on the training set constructed by SPLIT at the end of lap 2. The errors exhibit a continuous and smooth gradient, with adjacent regions showing high similarity. This regularity indicates that the low-dimensional features $\begin{bmatrix} \alpha_f, \alpha_r, T\end{bmatrix}^\text{T}$possess the capability to capture model deviations associated with the variable elements of the vehicle model.}
\label{figure: model error}
\setlength{\abovecaptionskip}{0.cm}
\end{figure*}

\begin{figure*}[!t]
    \centering
\includegraphics{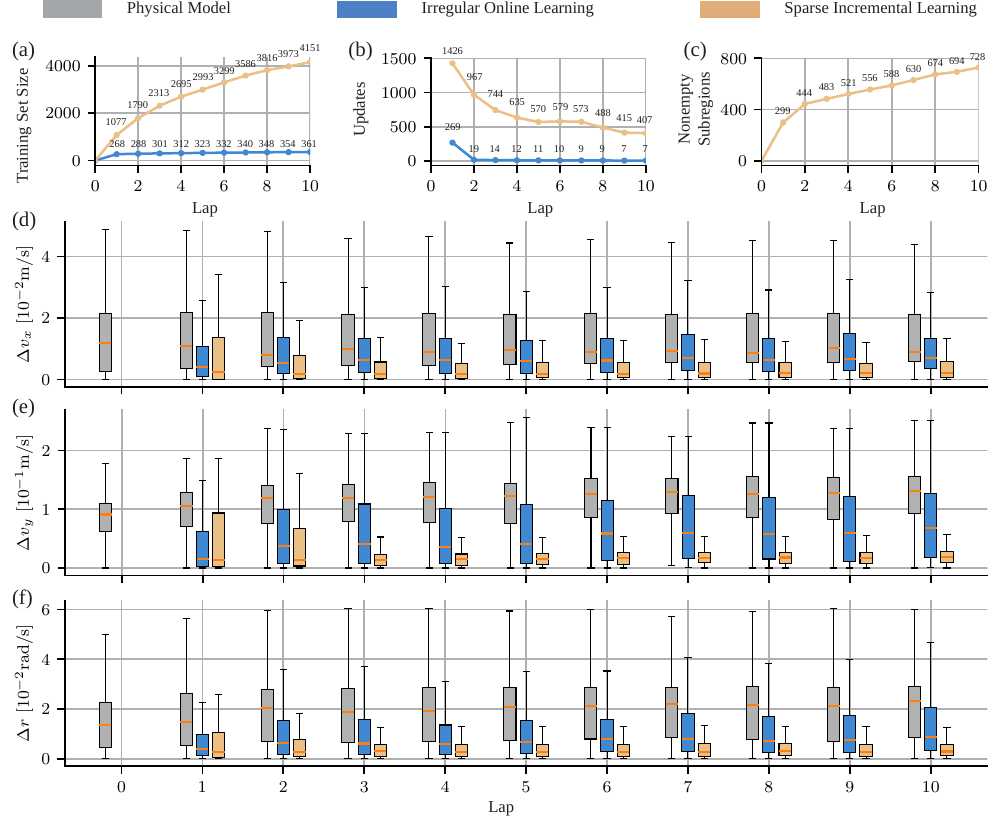}
\caption{Comparison of error dynamics learning with SPLIT and IOL. (a)–(c) Efficiency of online learning, showing training set size, number of updates per lap, and number of non-empty subregions. (d)–(f) Prediction performance on the racing data collected under the SPLIT-enhanced controller. Since IOL did not learn on this data, its predictions are generated using the training sets obtained at the end of each lap with the IOL-enhanced controller. SPLIT achieves far higher learning efficiency, with substantially larger training sets and more frequent updates per lap. While the physical model errors increase with racing aggressiveness, SPLIT consistently suppresses deviations to a sufficient low level, whereas IOL provides only limited improvement due to inefficient updates. The higher errors of SPLIT in the first lap arise from the initially empty training set.}
\label{figure: impact of the sparse incremental learning}
\setlength{\abovecaptionskip}{0.cm}
\end{figure*}

The first reason for SPLIT to have a stronger enhancement on controller performance is that the residual model in Section \ref{subsection: residual model} only compensates for deviations in the variable elements of the nominal model, reducing the dimensionality of input features. This dimensionality reduction allows a training set of the same size to cover a wider range of vehicle operating conditions. However, it remains questionable whether this simplification is reasonable and whether it may lead to incorrect model error predictions.

To answer this question, we compare the effects of two feature selections on the residual model efficacy in compensating for model deviations:

\begin{itemize}
\item $\bm{z}_1=[v_x,v_y,r,T,\delta]^T$, the dynamic part of vehicle states, which compensates all potential model deviations \cite{hewing2018cautious, kabzan2019learning, hewing2019cautious, jiang2021high, jiang2023fast, zhao2024learning}.
\item $\bm{z}_2=[\alpha_f,\alpha_r,T]^T$, which compensates deviations only from variable elements.
\end{itemize}
To eliminate the confounding effect of varying training set sizes, we use the marginal gain metric defined in (\ref{equation: independence measurement}) to select the most informative samples from lap 0 data, construct training sets of size 100 for different features and develop corresponding residual models.
The compensation efficacy of each residual model is quantified by comparing the prediction errors during lap 0: \(||\bm{e}_{g}|| = ||\bm{x}_{k+1} - \bm{f}(\bm{x}_k, \bm{u}_k) - \bm{B}_d \bm{g}(\bm{x}_k, \bm{u}_k)||\). 
The prediction error of nominal model is computed as \(||\bm{e}_{n}|| = ||\bm{x}_{k+1} - \bm{f}(\bm{x}_k, \bm{u}_k)||\), providing a reference for evaluating improvements in model fidelity.
Figure \ref{figure: compensation for model error in variable elements} illustrates the prediction errors of different feature selections across state components. 
The results indicate that the prediction errors of $\bm{z}_1$ are smaller than those of $\bm{z}_2$ and the model deviations compensated by feature $\bm{z}_2$ yields only limited improvement over the physical model. 
This phenomenon is attributed to the limited coverage of high-dimensional feature spaces by limited training sets. Dimensionality reduction effectively expands the range of vehicle operating conditions that the residual model can compensate.


Figure \ref{figure: model error} further illustrates the distribution of the physical model error in velocity states within the feature space.
The model errors exhibit a continuous and smooth gradient trend, with the errors in adjacent regions within the feature space being highly similar.
This regularity confirms that the three features in $\bm{z}_2$ can accurately capture model deviations from the variable elements of the vehicle model.


\subsection{Impact of the Sparse Incremental Learning}
\label{subsection: impact of the sparse incremental learning}

\begin{figure}[!t]
    \centering
\includegraphics{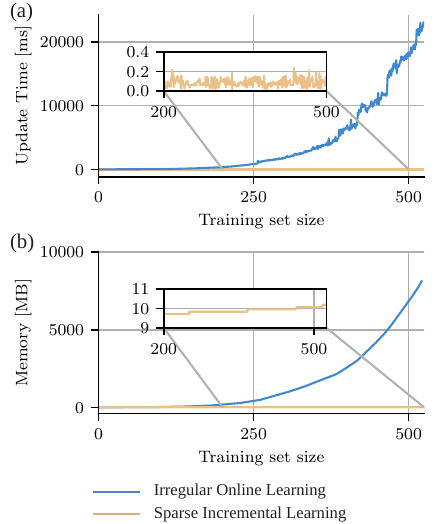}
\caption{Comparison of SPLIT and IOL under different training set sizes: (a) update time and (b) memory usage. SPLIT maintains both metrics at minimal levels, making it suitable for in-vehicle controllers, whereas IOL becomes impractical for real-time control once the training set grows even moderately.}
\label{figure: update time}
\setlength{\abovecaptionskip}{0.cm}
\end{figure}

The second reason is the Sparse Incremental Learning mechanism, which streamlines the evaluation of marginal coverage gain and substantially enhances the efficiency of updating the training set. To quantitatively assess its benefits, a comparative analysis is conducted between SPLIT and IOL, which directly updates the entire training set.

Figure \ref{figure: impact of the sparse incremental learning} (a)-(c) illustrate the update frequency and training set size achieved by SPLIT and IOL throughout the racing task, and the number of non-empty subregions within the valid region. After just one lap, SPLIT performs 1426 times updates and constructs a training set of 1077 samples. In contrast, IOL only yields a training set of merely 268 samples. 
As the training set size increases, the learning efficiency of IOL declines significantly, resulting in only 361 samples by the end of lap 10. In contrast, SPLIT continuously expands its training set, ultimately reaching 4151 samples covering 728 subregions within the valid region, corresponding to 60\% of the total.

This significant disparity arises from the substantial reduction in computational complexity per update achieved by SPLIT. As shown in Fig. \ref{figure: update time}, SPLIT maintains a consistent update time below 0.2 ms, which is significantly lower than the 50 ms control sampling interval and remains invariant to the training set size. This enables the use of every measurement frame during operation. In contrast, the update time of IOL increases rapidly to over 20 s as the training set expands, severely limiting the efficiency of online learning. The high priority assigned to the control thread further restricts available computational resources, resulting in additional delays in training set updates. Additionally, SPLIT has a far lower memory requirement than IOL, making it suitable for deployment on in-vehicle hardware.

Figure \ref{figure: impact of the sparse incremental learning} (d)-(f) further compare the real-time compensation of model errors achieved by SPLIT and IOL during racing. 
The prediction errors of the physical model increase with the growing aggressiveness of the racing. Although IOL partially improves model accuracy, its compensation capacity is limited due to the diminishing learning efficiency. In contrast, SPLIT successfully maintains model errors at a consistently low and stable level. The slightly higher errors observed during the first two laps are attributed to the initially empty training set, but the compensation performance rapidly improves as real-time measurements are incorporated into the training set.

\subsection{Impact of the Sparse Gaussian Process}
\label{subsection: impact of the sparse gaussian process}

\begin{figure*}[!t]
    \centering
\includegraphics{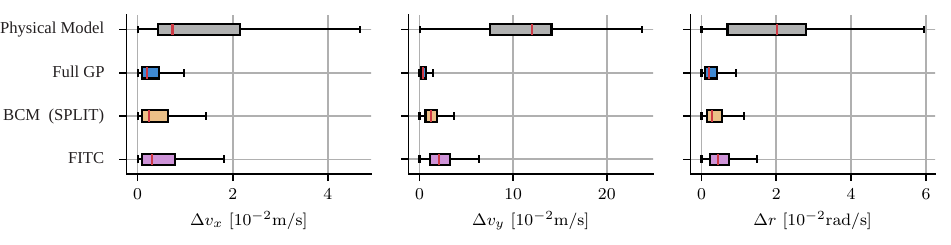}
\caption{Comparison of different online evaluation methods with respect to model accuracy. The evaluation uses data from the SPLIT-enhanced controller in lap 2, with training sets constructed at the end of lap 2. Although BCM does not reach the theoretical optimum of the full GP, the gap is minor and its performance remains superior to FITC.}
\label{figure: impact of the sparse gaussian process}
\setlength{\abovecaptionskip}{0.cm}
\end{figure*}

\begin{figure}[!t]
    \centering
\includegraphics{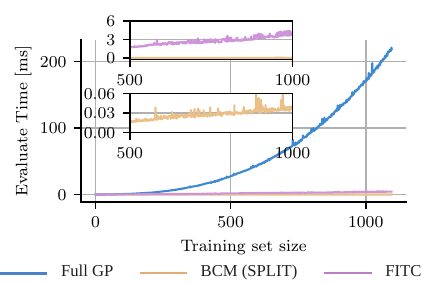}
\caption{Comparison of online evaluation time. Both BCM and FITC significantly reduce the evaluation time of GP, with BCM showing a more pronounced advantage. This efficiency gain arises not only from simplifying kernel inversion through multiple low-dimensional GPs in parallel, but also from reusing kernel matrices obtained during the online learning.}
\label{figure: evaluate time}
\setlength{\abovecaptionskip}{0.cm}
\end{figure}

The third reason is the sparse GP introduced in Section \ref{subsection: sparse gaussian process}, which significantly reduces the computational complexity associated with online evaluation. This reduction is critical for enabling real-time evaluation over the large-scale training set constructed by SPLIT. To rigorously assess the impact of sparsification on online evaluation, we compare the predictive performance of several GP evaluation methods.
\begin{itemize}
\item Full GP. Evaluation is performed directly using the formulation in (\ref{equation: gp prediction}), without incorporating sparsification.
\item FITC \cite{snelson2005sparse}. Evaluation with pseudo-input-based sparsification, where ten future states at 100 ms intervals are selected as pseudo-inputs as in \cite{hewing2018cautious, hewing2019cautious}.
\item BCM. Evaluation follows the BCM formulation as defined in (\ref{equation: bcm prediction}).
\end{itemize}
The evaluation is based on measurements acquired in lap 1 of SPLIT-enhanced controller, with corresponding training set comprising samples collected by SPLIT at the end of lap 1.

Figure \ref{figure: impact of the sparse gaussian process} presents the open-loop prediction results. The Full GP serves as the reference, representing the best achievable model accuracy given the current training set. In comparison, BCM introduces slight deviations across all states, resulting in a minor reduction in prediction accuracy, yet it still outperforms FITC. Given the model errors of the original physical model, the marginal degradation introduced by BCM is negligible and has limited impact on control performance. However, the reduction in computational complexity brought by BCM is of significant practical value.

Figure \ref{figure: evaluate time} further illustrates the computation time to perform a single evaluation under varying training set sizes. The computational cost of Full GP increases exponentially with the training set size, rapidly exceeding the sampling interval of controller and becoming unsuitable for real-time applications. In contrast, BCM maintains the evaluation time of only 0.04 ms even with a training set of size 1000, and its computational complexity scales linearly with the training set size. Compared with FITC, BCM demonstrates clear advantages in both prediction accuracy and computational efficiency.

The efficiency is primarily facilitated by performing multiple low-dimensional GP evaluations instead of full kernel matrix inversion.
Parallel processing of each small-scale GP further decreases computational time for individual evaluations.
Additionally, the seamless integration of SPLIT with BCM allows the gram matrix \(\bm{K}_i\) and its inverse \(\bm{K}_i^{-1}\), computed during online learning for each subset \(\mathcal{D}_i\), to be directly reused in BCM evaluation, eliminating redundant computations.



\section{Real-World Experiments}
\label{section: real-world experiments}

In this section, we deploy SPLIT on a real vehicle platform to evaluate its effectiveness in real-world driving conditions. Four scenarios are considered as follows:

\begin{itemize}
    \item \textbf{Autonomous Racing}. Evaluates the online learning efficacy using streaming data.
    \item \textbf{Double Lane Change}. Evaluates the ability of rapid adaptation to model deviation.
    \item \textbf{Collision Avoidance}. Evaluates generalization to unseen lateral control scenario.
    \item \textbf{Emergency Brake and Avoidance}. Evaluates generalization to unseen coupled longitudinal–lateral control scenario, which is safety-critical..
\end{itemize}
Clips of all the experiments could be found in the Supplementary Material.

\subsection{Experimental Platform and Configuration}
\label{subsection: experimental platform and configuration}

\begin{figure}[!t]
    \centering
\includegraphics[]{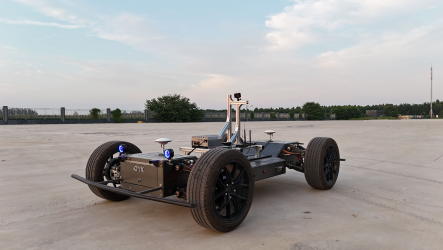}
\caption{Experimental platform. A four-wheel independently driven electric chassis equipped with an integrated navigation system for state measurement and an industrial computer as the control unit.}
\label{figure: pix}
\setlength{\abovecaptionskip}{0.cm}
\end{figure}


In real-world experiments, we employ an electric chassis as the experimental platform, as illustrated in Fig. \ref{figure: pix}. 
This chassis enables direct actuation of drive torque, front-wheel steering angle, and braking pressure through CAN interface, with a 20 ms control interval.
The tire parameters are identified through tire experiments. Due to mismatches between the experimental conditions and real-world road surfaces, deviations between the model and actual tire dynamics are inevitably introduced.
We employ the SPLIT framework to perform online compensation for these structured deviations.

The chassis is equipped with a integrated navigation system comprises a differential GPS and inertial measurement unit (IMU) to measure vehicle position, velocity, yaw angle, yaw rate and acceleration at a sampling rate of 100 Hz. The front wheel steering angle and driving torque are acquired via CAN messages at 100 Hz.
Consistent with the simulation setup, the control algorithm is implemented in C++ and deployed on a Nuvo-9160 industrial computing platform equipped with an Intel i9-13900 CPU. Control commands are transmitted to the vehicle via a CANNET equipment, which converts UDP signals into CAN messages.

All the experiments are conducted on a homogeneous cement road surface. The experimental details are elaborated in the following.

\subsection{Scenario \uppercase\expandafter{\romannumeral1}: Autonomous Racing}
\label{subsection: scenario: autonomous racing}

\begin{figure}[!t]
    \centering
\includegraphics[]{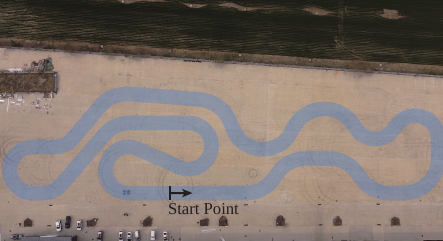}
\caption{Illustration of the race track in real-vehicle experimental scenario \uppercase\expandafter{\romannumeral1}. The track has a length of 579.1 m, consisting of multiple aggressive curves and straight segments, with a homogeneous concrete surface. }
\label{figure: race track}
\setlength{\abovecaptionskip}{0.cm}
\end{figure}

\subsubsection{Experimental Setup}
In the scenario \uppercase\expandafter{\romannumeral1}, we employ the autonomous racing task to evaluate the online learning efficacy of SPLIT in the real-world. The race track is shown in Fig. \ref{figure: race track} and has a total length of 579.1 m. To ensure operational safety, the maximum driving torque is limited to 252 N·m, and the maximum vehicle velocity is set as 15 m/s. At each control sampling interval, SPLIT updates the model in real time using the most recent measurements. The updated model is then used to compute the control input.
The experiment comprises 10 laps, with an additional lap conducted using the nominal controller as the baseline, denoted as lap 0.
\subsubsection{Controller}

Consistent with simulation experiments, the controller employed is the MPCC controller in (\ref{equation: mpcc formulation}), which operates at a sampling interval of 40 ms. All other configurations remain identical with the simulation experiments.

\begin{figure*}[!t]
    \centering
\includegraphics{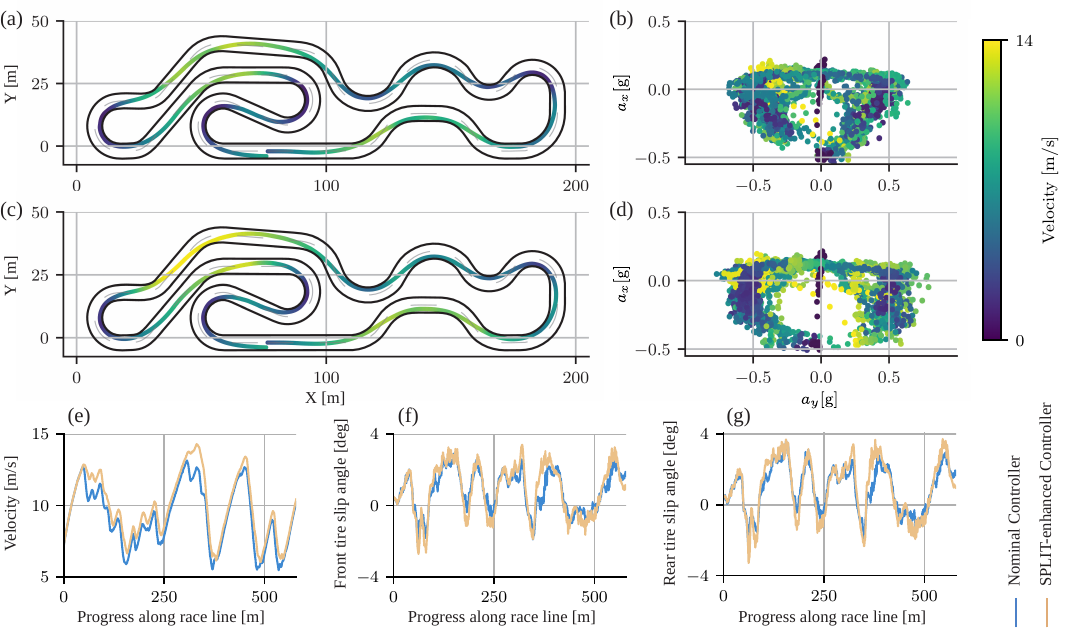}
\caption{Experimental results of Scenario \uppercase\expandafter{\romannumeral1}: Autonomous Racing. Comparison between the nominal controller and the SPLIT-enhanced controller on the best lap: (a)–(b) trajectory and GG-diagram with the nominal controller; (c)–(d) trajectory and GG-diagram with the SPLIT-enhanced controller; (e) velocity profile; (f)–(g) front and rear tire slip angles during racing. SPLIT improves the capability of controller to handle vehicle dynamics, enabling more aggressive driving and higher velocity across almost every track segments. Slightly different from simulation, the strong coupling between lateral and longitudinal tire forces is more prominent and limits the nominal controller under combined braking and cornering conditions. In contrast, SPLIT captures this coupling, leading to a broader GG-diagram with a distinctly different shape.}
\label{figure: scenario 1}
\setlength{\abovecaptionskip}{0.cm}
\end{figure*}

\subsubsection{Results}

\begin{table}[!t]
\caption{\textbf{Experimental results of Scenario I: Autonomous Racing}}
\label{table: autonomous racing results}
\centering
\begin{tabular}{ccccccc}
\toprule
\multirow{2}{*}{Lap} & Time & $\Vert a_y \Vert_{\text{max}}$ & Data & Training & Nonempty\\
& [s] & [g] & Updates & set size & subregion\\
\midrule
0&60.12&0.68&-&-&- \\
1&58.36&0.69&700&443&124 \\
2&57.80&0.71&185&607&160 \\
3&57.56&0.74&186&718&177 \\
4&57.20&0.75&164&792&188 \\
5&56.68&0.74&172&914&202 \\
6&56.08&0.77&127&990&210 \\
7&56.24&0.76&117&1075&226 \\
8&55.84&0.78&111&1116&233 \\
9&56.04&0.80&96&1197&243 \\
10&55.72&0.79&85&1254&261 \\
\bottomrule
\end{tabular}
\end{table}

Table \ref{table: autonomous racing results} summarizes key performance metrics during the entire autonomous racing and Fig. \ref{figure: scenario 1} further supplements the detailed comparison between the best lap achieved with SPLIT-enhanced controller and the lap 0 obtained using the nominal controller. Consistent with the simulation, SPLIT incrementally expands the training set using real-time measurements, enhancing the capability of the controller to handle actual vehicle dynamics and enabling more effective utilization of tire adhesion, resulting in a reduction in lap time from 60.12 s to 55.72 s and a notable increase in the maximum lateral acceleration from 0.68 g to 0.80 g. 

Slightly different from simulation experiments, the strong coupling between lateral and longitudinal tire forces is more prominent. Such interaction is not explicitly modeled in the nominal vehicle model, leading to deviations between the computed and actual tire forces, which in turn limits the controller ability to handle the vehicle dynamics. The GG-diagram of lap 0 in Fig. \ref{figure: scenario 1} (b) shows that the data are insufficiently spread along the lateral acceleration axis, and the distribution narrows rapidly as the magnitude of longitudinal acceleration increases. This indicates that the nominal controller fails to fully exploit the available tire adhesion, particularly under combined braking and steering conditions.

In contrast, SPLIT effectively compensates for such mathematically intractable unmodeled deviations. The GG-diagram of the best lap in Fig. \ref{figure: scenario 1} (d) demonstrates a wider lateral acceleration distribution and a larger central void region, suggesting that the vehicle consistently operates under more aggressive conditions. Additionally, the lateral dispersion of the GG-diagram is more notably improved along the negative vertical axis, confirming that the controller can more precisely coordinate braking and steering during racing.

\begin{figure*}[!t]
    \centering
\includegraphics{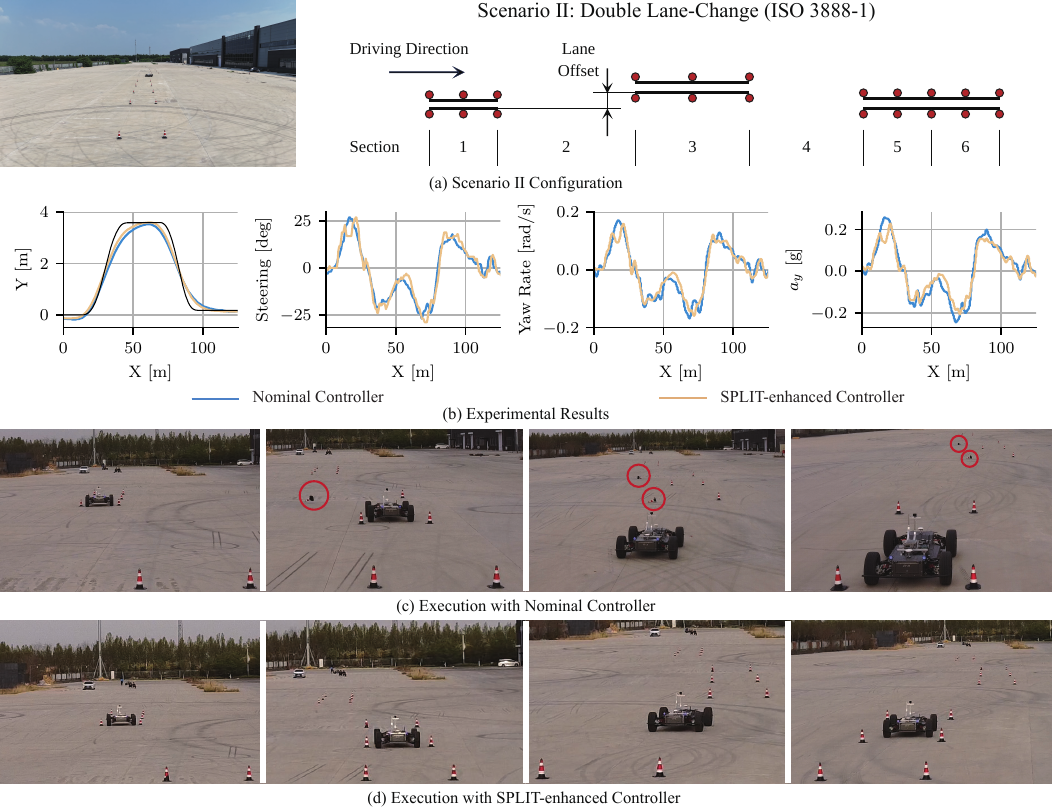}
\caption{Configuration and experimental results of Scenario \uppercase\expandafter{\romannumeral2}: Double Lane Change (ISO 3888-1). The experiment demonstrates the rapid adaptation capability of SPLIT to vehicle dynamics deviations. Due to discrepancies between the physical model and real vehicle dynamics, the nominal controller collides with cones when entering and exiting section 3, resulting in experimental failure. In contrast, SPLIT only leverages the data segment of less than 10 s from the nominal controller to quickly adapt to this operating condition and enables collision-free completion of the maneuver. Red circles indicate the cones knocked down in the experiment.}
\label{figure: iso 3888-1}
\setlength{\abovecaptionskip}{0.cm}
\end{figure*}

\subsection{Scenario \uppercase\expandafter{\romannumeral2}: Double Lane Change (ISO 3888-1)}
\label{subsection: scenario: double lane}

\begin{table}[!t]
\caption{\textbf{Scenario Configuration}}
\label{table: scenario setup}
\centering
\begin{tabular}{c|ccc|ccc}
\toprule
\multicolumn{1}{c|}{\multirow{3}{*}{Section}} & \multicolumn{3}{c|}{ISO 3888-1} & \multicolumn{3}{c}{ISO 3888-2} \\ 
\cmidrule(lr){2-4} \cmidrule(l){5-7} 
\multicolumn{1}{c|}{} & \multirow{2}{*}{Length} & \multirow{2}{*}{Width} & Lane  & \multirow{2}{*}{Length} & \multirow{2}{*}{Width} & Lane\\
\multicolumn{1}{c|}{}&  &  & Offset &  &  & Offset\\
\midrule
1&15&2.01& - & 12 &2.01&- \\
2&30&  - & - &13.5&  - &- \\
3&25&2.17&3.5& 11 &2.60&3.01 \\
4&25&  - & - &12.5& -  &- \\
5&15&2.33& - & 12 &3.00&- \\
6&15&2.33& - &  - &  - &- \\
\bottomrule
\addlinespace[2pt] 
\multicolumn{7}{r}{Dimensions in metres} 
\end{tabular}
\end{table}

\subsubsection{Experimental Setup}

In Scenario \uppercase\expandafter{\romannumeral2}, the double lane-change maneuver in ISO 3888-1 \cite{ISO3888-1} is employed to evaluate the ability of SPLIT to rapidly adapt to model deviations. The track configuration is illustrated in Fig. \ref{figure: iso 3888-1} (a), and the lengths and widths of each section are summarized in Table \ref{table: autonomous racing results}. The experiment is conducted at the highest feasible velocity of 14 m/s. Additional experimental details can be found in ISO 3888-1 \cite{ISO3888-1}. The experiment consists of two executions: in the first, the nominal controller operates while real-time measurements are fed into SPLIT to update the residual model online; in the second, the SPLIT-enhanced controller carries out the maneuver to demonstrate its improved performance.


\begin{figure*}[!t]
    \centering
\includegraphics{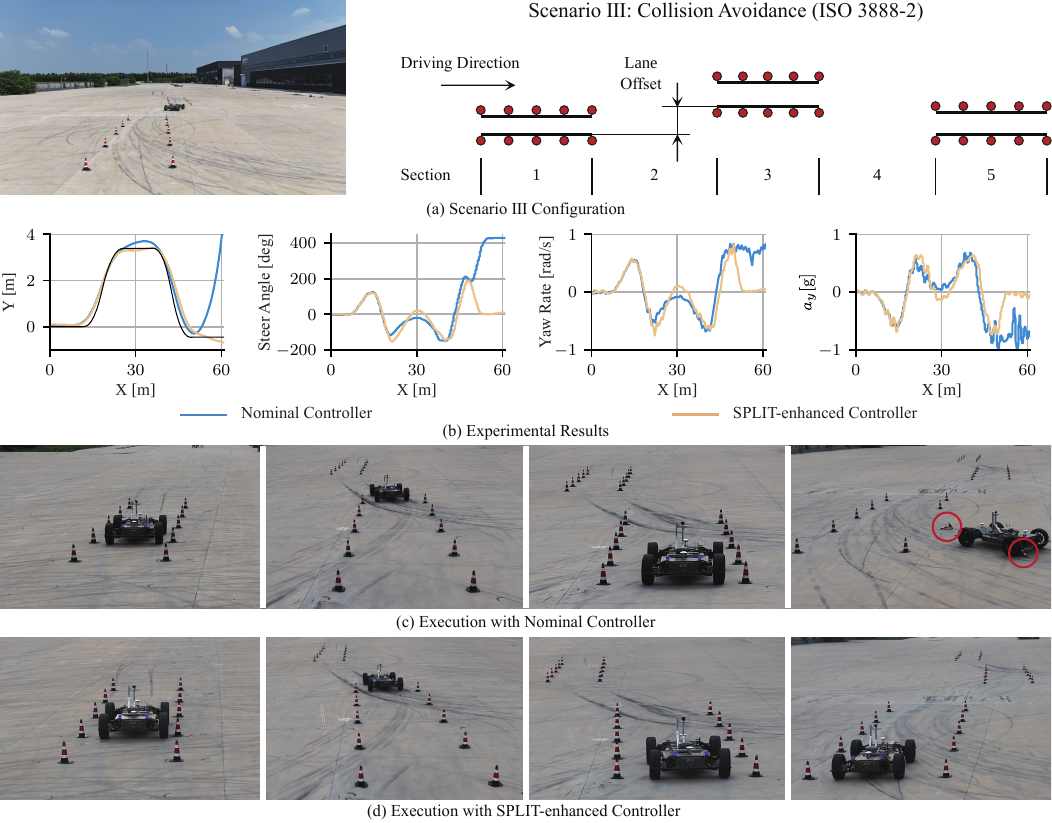}
\caption{Configuration and experimental results of Scenario \uppercase\expandafter{\romannumeral3}: Collision Avoidance (ISO 3888-2). The experiment demonstrates the robust generalization capability of SPLIT to unseen driving conditions. In this scenario, the nominal controller deviates significantly from the reference trajectory at the end of the second lane change, collides with cones, and leads to experimental failure. In contrast, the SPLIT-enhanced controller satisfies trajectory constraints, completes the maneuver without collisions, and improves tracking accuracy. The training set used by SPLIT is constructed from less than 60 s of data collected at the end of lap 1 in Scenario \uppercase\expandafter{\romannumeral1}. Red circles indicate the cones knocked down in the experiment.}
\label{figure: iso 3888-2}
\setlength{\abovecaptionskip}{0.cm}
\end{figure*}

\subsubsection{Controller}
The longitudinal motion of the vehicle is regulated by a simple PID controller to maintain the longitudinal velocity at the predefined reference value. The lateral motion is governed by the MPC controller introduced in Section \ref{subsection: model predictive control}, with a prediction horizon of 50 steps and a control sampling interval of 40 ms. 

\subsubsection{Results}
The experimental results of ISO 3888-1 are presented in Fig. \ref{figure: iso 3888-1}. 
Due to inaccuracies in the vehicle model, the vehicle controlled by nominal controller collides with cones marking the trajectory boundaries when entering and exiting section 3, as shown in Fig. \ref{figure: iso 3888-1} (c). During this run, SPLIT continuously receives the real-time measurements and performs online updates of residual model using less than 10 s of trajectory data. Despite the very limited data length, SPLIT effectively compensates for local model deviations within the corresponding feature space. When the maneuver is repeated under the same conditions, the SPLIT-enhanced controller successfully completes the test without contacting any cones, as illustrated in Fig. \ref{figure: iso 3888-1} (d). Fig. \ref{figure: iso 3888-1} (b) presents a comparison of the two experimental executions, showing that the closed-loop trajectory obtained with the SPLIT-enhanced controller aligns more closely with the reference trajectory.



\subsection{Scenario \uppercase\expandafter{\romannumeral3}: Collision Avoidance (ISO 3888-2)}
\label{subsection: scenario: collision avoidance}

\subsubsection{Experimental Setup}

In Scenario \uppercase\expandafter{\romannumeral3}, the collision avoidance maneuver in ISO 3888-2 \cite{ISO3888-2} is employed to evaluate the generalization capability of SPLIT in lateral control scenario, namely its ability to compensate for model deviations in a novel driving scenario. 
The configuration of the collision avoidance track is illustrated in Fig. \ref{figure: iso 3888-2}, and the lengths and widths of each section are summarized in Table \ref{table: scenario setup}. The experiment is conducted at a velocity of 10 m/s, and additional configuration details can be found in ISO 3888-2 \cite{ISO3888-2}. 
Two executions are performed: the first using the nominal controller, and the second using the SPLIT-enhanced controller. It is important to note that the training set utilized by SPLIT in this scenario originates from Scenario \uppercase\expandafter{\romannumeral1}, specifically from the training set obtained after lap 1 with the SPLIT-enhanced controller.
Consequently, the collision avoidance maneuver constitutes an entirely unseen scenario for SPLIT.

\begin{figure*}[!t]
    \centering
\includegraphics{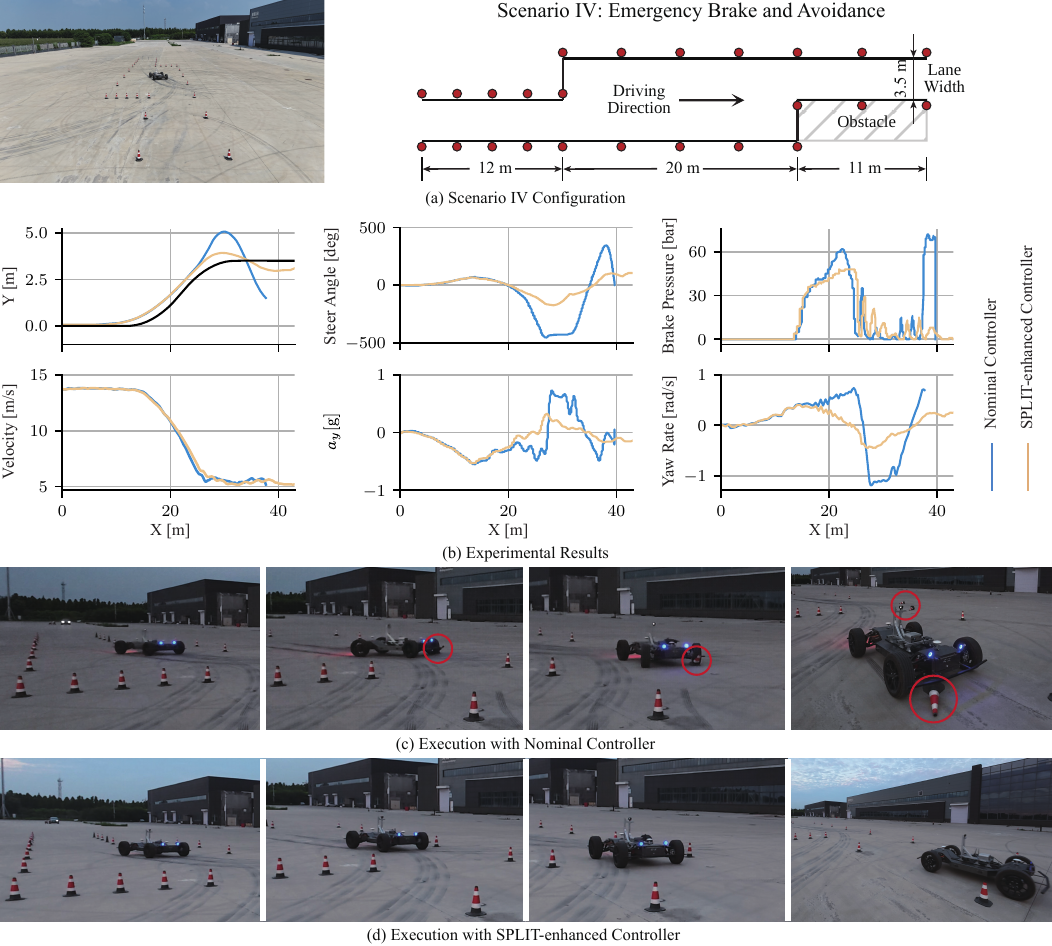}
\caption{Configuration and experimental results of Scenario \uppercase\expandafter{\romannumeral4}: Emergency brake and avoidance. The experiment demonstrates the robust generalization capability of SPLIT to unseen driving conditions. In this scenario, the nominal controller fails to handle the coupling between longitudinal and lateral tire forces, causing the rear tires force to exceed the road adhesion limit, which leads to vehicle instability, deviation from the predefined trajectory, and cone collisions. In contrast, the SPLIT-enhanced controller effectively coordinates longitudinal and lateral commands, maintains steering capability, and completes the maneuver without collisions. The training set used by SPLIT is constructed from less than 60 s of data collected at the end of lap 1 in Scenario \uppercase\expandafter{\romannumeral1}.}
\label{figure: Emergency Brake and Avoidance}
\setlength{\abovecaptionskip}{0.cm}
\end{figure*}

\subsubsection{Controller}
The controller configuration in Scenario \uppercase\expandafter{\romannumeral3} remains identical to that in Scenario \uppercase\expandafter{\romannumeral2}. Given the more aggressive steering maneuvers, the trajectory boundaries are additionally modeled as soft constraints within the controller.

\subsubsection{Results}
The experimental results of Scenario \uppercase\expandafter{\romannumeral3} are shown in Fig. \ref{figure: iso 3888-2}. During the execution with the nominal controller, the controller fails near the end of the second lane change, leading to divergence of the closed-loop trajectory, a severe deviation from the reference path, and ultimately cone collisions that cause the experiment to fail. In contrast, the SPLIT-enhanced controller maintains the vehicle states within reasonable domains, ensuring stability while satisfying the trajectory constraints. Moreover, the SPLIT-enhanced controller achieves substantially better tracking accuracy than the nominal controller, with its trajectory nearly coinciding with the reference in section 3. These results demonstrate that SPLIT exhibits strong generalization capability, effectively compensates for model deviations in unseen scenarios, and enhances overall controller performance.

\subsection{Scenario \uppercase\expandafter{\romannumeral4}: Emergency Brake and Avoidance}
\label{subsection: scenario: emergency brake and collision}

\subsubsection{Experimental Setup}

In scenario \uppercase\expandafter{\romannumeral4}, the emergency brake and avoidance maneuver is employed to further evaluate the ability of SPLIT to compensate for model deviations in unseen coupled longitudinal-lateral scenario. The configuration of this scenario, illustrated in Fig. \ref{figure: Emergency Brake and Avoidance}, simulates an autonomous vehicle traveling at high velocity when suddenly encountering an obstacle and performing an emergency avoidance maneuver. 
The experimental vehicle enters the predefined lane at an initial speed of 14 m/s and, within a 20 m distance, simultaneously executes braking and steering: the longitudinal velocity is reduced to 5 m/s while the lateral position is shifted to the adjacent lane to avoid the obstacle. 
The experiment is executed once with the nominal controller and once with the SPLIT-enhanced controller. The training set used by SPLIT is the same as that in Scenario \uppercase\expandafter{\romannumeral3}, constructed from less than 60 s of racing data. Since the emergency brake and avoidance scenario rarely occurs in normal driving, this experiment serves to emulate how SPLIT, trained solely on regular driving data, can enhance the safety of autonomous vehicles when confronted with entirely new and safety-critical situations.

\subsubsection{Controller}

Both the longitudinal and lateral motions of vehicle are jointly controlled by MPC controller formulated in (\ref{equation: mpcc formulation}). The prediction horizon is set to 25 steps with a sampling interval of 40 ms. The trajectory boundaries are modeled as soft constraints and embedded into the controller.

\subsubsection{Results}
The experimental results of Scenario \uppercase\expandafter{\romannumeral4} are presented in Fig. \ref{figure: Emergency Brake and Avoidance}. Under the nominal controller, the physical model fails to accurately capture the coupling between longitudinal and lateral tire forces. As the vehicle simultaneously performs steering and braking, the combined tire force exceeds the road adhesion limit, causing severe slip. Consequently, the vehicle violates the trajectory constraints, collides with two cones on one side, and during corrective actions further strikes cones on the opposite side. For safety reasons, the experiment is manually terminated.
In contrast, the SPLIT-enhanced controller effectively accounts for the influence of longitudinal tire force on lateral force. During the lane-change maneuver, it avoids excessively increasing the steering angle in pursuit of higher tracking accuracy; instead, it adopts a relatively smaller steering input to preserve steering capability and prevent vehicle loss of control. This strategy achieves accurate coordination of longitudinal and lateral dynamics, enabling the vehicle to complete the maneuver without collisions.
More detailed experimental data are shown in Fig. \ref{figure: Emergency Brake and Avoidance} (b).
These results validate the capability of SPLIT to enhance control performance in unseen scenarios by online learning from regular driving data.

\section{Conclusion}
\label{section: conclusion}

In summary, this article proposes SPLIT, a hybrid control-oriented modeling framework that leverages streaming data to learn model error online across the entire vehicle performance envelope. 
SPLIT decomposes the vehicle model so that the feature dimension of the GP-based residual model is reduced from 5 to 3, which exponentially decreases the required training set size.
By explicitly defining and partitioning the valid region into \(K\) subregions, the model error learning task is transformed into data updates within corresponding subregion, reducing the computational complexity of marginal gain evaluation from \(\mathcal{O}(K^3M^3)\) to \(\mathcal{O}(M^3)\) and the number of required evaluations from \(KM\) to \(M\). This formulation achieves a single update time below 0.2 ms and a total memory usage of about 10 MB.
The segmented subsets are further utilized to sparsify the online GP evaluation, reducing the computational complexity from \(\mathcal{O}(K^3M^3)\) to \(\mathcal{O}(KM^3)\), and the evaluation time to well below the controller sampling interval.

Simulation and real-world experiments demonstrate that SPLIT maintains vehicle model discrepancies at a consistently low and stable level, and improve control performance online.
SPLIT also exhibits rapid adaptation to model deviations, achieving improvements in control performance with only about 10 s of data from the corresponding scenarios.
Moreover, SPLIT exhibits robust generalization to previously unseen scenarios. 
We believe that SPLIT provides an important stepping stone toward the deployment of GP-based residual models in autonomous driving control and can be readily transferred to wheeled robots with similar structures.



\bibliography{references}

@INPROCEEDINGS{li2025learning,
  author={Li, Yaoyu and Huang, Chaosheng and Yang, Dongsheng and Liu, Wenbo and Li, Jun},
  booktitle={2025 IEEE International Conference on Robotics and Automation (ICRA)}, 
  title={Learning Based MPC for Autonomous Driving Using a Low Dimensional Residual Model}, 
  year={2025},
  volume={},
  number={},
  pages={10958-10964},
  doi={10.1109/ICRA55743.2025.11128642}}

@ARTICLE{askari2025Model,
  author={Askari, Iman and Vaziri, Ali and Tu, Xuemin and Zeng, Shen and Fang, Huazhen},
  journal={IEEE Transactions on Robotics}, 
  title={Model Predictive Inferential Control of Neural State-Space Models for Autonomous Vehicle Motion Planning}, 
  year={2025},
  volume={41},
  number={},
  pages={3202-3222},
  doi={10.1109/TRO.2025.3566198}}

@article{stano2023model,
title = {Model predictive path tracking control for automated road vehicles: A review},
journal = {Annual Reviews in Control},
volume = {55},
pages = {194-236},
year = {2023},
issn = {1367-5788},
doi = {https://doi.org/10.1016/j.arcontrol.2022.11.001},
author = {P. Stano and U. Montanaro and D. Tavernini and M. Tufo and G. Fiengo and L. Novella and A. Sorniotti}
}

@article{hewing2020learning,
title = {Learning-Based Model Predictive Control: Toward Safe Learning in Control},
journal = {Annual Review of Control, Robotics, and Autonomous Systems},
volume = {3},
pages = {269-296},
year = {2020},
issn = {2573-5144},
doi = {https://doi.org/10.1146/annurev-control-090419-075625},
author = {Hewing, Lukas and Wabersich, Kim P. and Menner, Marcel and Zeilinger, Melanie N.}
}

@ARTICLE{zhang2024survey,
  author={Zhang, Tantan and Sun, Yueshuo and Wang, Yazhou and Li, Bai and Tian, Yonglin and Wang, Fei-Yue},
  journal={IEEE Transactions on Intelligent Vehicles}, 
  title={A Survey of Vehicle Dynamics Modeling Methods for Autonomous Racing: Theoretical Models, Physical/Virtual Platforms, and Perspectives}, 
  year={2024},
  volume={9},
  number={3},
  pages={4312-4334},
  doi={10.1109/TIV.2024.3351131}}

@article{yang2013overview,
  title={An overview on vehicle dynamics},
  author={Yang, Shaopu and Lu, Yongjie and Li, Shaohua},
  journal={International Journal of Dynamics and Control},
  volume={1},
  pages={385--395},
  year={2013},
  publisher={Springer}
}

@article{kabzan2019learning,
  author={Kabzan, Juraj and Hewing, Lukas and Liniger, Alexander and Zeilinger, Melanie N.},
  journal={IEEE Robotics and Automation Letters}, 
  title={Learning-Based Model Predictive Control for Autonomous Racing}, 
  year={2019},
  volume={4},
  number={4},
  pages={3363-3370},
  doi={10.1109/LRA.2019.2926677}}

@INPROCEEDINGS{hewing2018cautious,
  author={Hewing, Lukas and Liniger, Alexander and Zeilinger, Melanie N.},
  booktitle={2018 European Control Conference (ECC)}, 
  title={Cautious NMPC with Gaussian Process Dynamics for Autonomous Miniature Race Cars}, 
  year={2018},
  volume={},
  number={},
  pages={1341-1348},
  doi={10.23919/ECC.2018.8550162}
}

@ARTICLE{hewing2019cautious,
  author={Hewing, Lukas and Kabzan, Juraj and Zeilinger, Melanie N.},
  journal={IEEE Transactions on Control Systems Technology}, 
  title={Cautious Model Predictive Control Using Gaussian Process Regression}, 
  year={2020},
  volume={28},
  number={6},
  pages={2736-2743},
  doi={10.1109/TCST.2019.2949757}
}

@INPROCEEDINGS{jiang2021high,
  author={Jiang, Shu and Wang, Yu and Lin, Weiman and Cao, Yu and Lin, Longtao and Miao, Jinghao and Luo, Qi},
  booktitle={2021 IEEE/RSJ International Conference on Intelligent Robots and Systems (IROS)}, 
  title={A High-accuracy Framework for Vehicle Dynamic Modeling in Autonomous Driving}, 
  year={2021},
  volume={},
  number={},
  pages={6680-6687},
  doi={10.1109/IROS51168.2021.9636861}
}

@article{armin2023Integrating,
title = {Integrating Machine Learning and Model Predictive Control for automotive applications: A review and future directions},
journal = {Engineering Applications of Artificial Intelligence},
volume = {120},
pages = {105878},
year = {2023},
issn = {0952-1976},
doi = {https://doi.org/10.1016/j.engappai.2023.105878},
author = {Armin Norouzi and Hamed Heidarifar and Hoseinali Borhan and Mahdi Shahbakhti and Charles Robert Koch}
}

@INPROCEEDINGS{ostafew2014learning,
  author={Ostafew, Chris J. and Schoellig, Angela P. and Barfoot, Timothy D.},
  booktitle={2014 IEEE International Conference on Robotics and Automation (ICRA)}, 
  title={Learning-based nonlinear model predictive control to improve vision-based mobile robot path-tracking in challenging outdoor environments}, 
  year={2014},
  volume={},
  number={},
  pages={4029-4036},
  doi={10.1109/ICRA.2014.6907444}
}

@article{ostafew2016learning,
  title={Learning-based nonlinear model predictive control to improve vision-based mobile robot path tracking},
  author={Ostafew, Chris J and Schoellig, Angela P and Barfoot, Timothy D and Collier, Jack},
  journal={Journal of Field Robotics},
  volume={33},
  number={1},
  pages={133--152},
  year={2016},
  publisher={Wiley Online Library}
}

@article{ostafew2016robust,
author = {Chris J. Ostafew and Angela P. Schoellig and Timothy D. Barfoot},
title ={Robust Constrained Learning-based NMPC enabling reliable mobile robot path tracking},
journal = {The International Journal of Robotics Research},
volume = {35},
number = {13},
pages = {1547-1563},
year = {2016},
doi = {10.1177/0278364916645661}
}

@INPROCEEDINGS{mckinnon2017learning,
  author={McKinnon, Christopher D. and Schoellig, Angela P.},
  booktitle={2017 IEEE International Conference on Robotics and Automation (ICRA)}, 
  title={Learning multimodal models for robot dynamics online with a mixture of Gaussian process experts}, 
  year={2017},
  volume={},
  number={},
  pages={322-328},
  doi={10.1109/ICRA.2017.7989041}}

@misc{scampicchio2025gaussian,
      title={Gaussian processes for dynamics learning in model predictive control}, 
      author={Anna Scampicchio and Elena Arcari and Amon Lahr and Melanie N. Zeilinger},
      year={2025},
      eprint={2502.02310},
      archivePrefix={arXiv},
      primaryClass={eess.SY},
      url={https://arxiv.org/abs/2502.02310}, 
}

@article{kapania2015design,
author = {Nitin R. Kapania and J. Christian Gerdes},
title = {Design of a feedback-feedforward steering controller for accurate path tracking and stability at the limits of handling},
journal = {Vehicle System Dynamics},
volume = {53},
number = {12},
pages = {1687--1704},
year = {2015},
publisher = {Taylor \& Francis},
doi = {10.1080/00423114.2015.1055279},
eprint = {https://doi.org/10.1080/00423114.2015.1055279}
}

@standard{ISO3888-1,
  title        = {Passenger cars — Test track for a severe lane-change manoeuvre — Part 1: Double lane-change},
  number       = {ISO 3888-1:2018},
  year         = {2018},
  institution  = {International Organization for Standardization},
  address      = {Geneva, Switzerland}
}

@standard{ISO3888-2,
  title        = {Passenger cars — Test track for a severe lane-change manoeuvre — Part 2: Obstacle avoidance},
  number       = {ISO 3888-2:2011},
  year         = {2011},
  institution  = {International Organization for Standardization},
  address      = {Geneva, Switzerland}
}

@inproceedings{zhao2024learning,
  title={Learning Residual Model of Model Predictive Control via Random Forests for Autonomous Driving},
  author={Zhao, Kang and Xue, Jianru and Meng, Xiangning and Li, Gengxin and Wu, Mengsen},
  booktitle={2024 IEEE 27th International Conference on Intelligent Transportation Systems (ITSC)},
  pages={755--761},
  year={2024},
  organization={IEEE}
}

@article{cao2024intelligent,
  title={Intelligent vehicle trajectory tracking control based on physics-informed neural network dynamics model},
  author={Cao, Xiuchen and Cai, Yingfeng and Li, Yicheng and Xiaoqiang, Sun and Chen, Long and Wang, Hai},
  journal={Proceedings of the Institution of Mechanical Engineers, Part D: Journal of Automobile Engineering},
  pages={09544070241244858},
  year={2024},
  publisher={SAGE Publications Sage UK: London, England}
}

@article{yang2024trajectory,
  title={Trajectory tracking control of autonomous vehicles based on Lagrangian neural network dynamics model},
  author={Yang, Wei and Cai, Yingfeng and Sun, Xiaoqiang and He, Youguo and Yuan, Chaochun and Wang, Hai and Chen, Long},
  journal={Proceedings of the Institution of Mechanical Engineers, Part D: Journal of Automobile Engineering},
  volume={238},
  number={12},
  pages={3483--3498},
  year={2024},
  publisher={SAGE Publications Sage UK: London, England}
}

@article{chrosniak2024deep,
  title={Deep dynamics: Vehicle dynamics modeling with a physics-constrained neural network for autonomous racing},
  author={Chrosniak, John and Ning, Jingyun and Behl, Madhur},
  journal={IEEE Robotics and Automation Letters},
  year={2024},
  publisher={IEEE}
}

@inproceedings{kim2022physics,
  title={Physics embedded neural network vehicle model and applications in risk-aware autonomous driving using latent features},
  author={Kim, Taekyung and Lee, Hojin and Lee, Wonsuk},
  booktitle={2022 IEEE/RSJ International Conference on Intelligent Robots and Systems (IROS)},
  pages={4182--4189},
  year={2022},
  organization={IEEE}
}

@article{kim2022toast,
  title={Toast: Trajectory optimization and simultaneous tracking using shared neural network dynamics},
  author={Kim, Taekyung and Lee, Hojin and Hong, Seongil and Lee, Wonsuk},
  journal={IEEE Robotics and Automation Letters},
  volume={7},
  number={4},
  pages={9747--9754},
  year={2022},
  publisher={IEEE}
}

@article{davydov2025first,
  title={First, Learn What You Don't Know: Active Information Gathering for Driving At the Limits of Handling},
  author={Davydov, Alexander and Djeumou, Franck and Greiff, Marcus and Suminaka, Makoto and Thompson, Michael and Subosits, John and Lew, Thomas},
  journal={IEEE Robotics and Automation Letters},
  year={2025},
  publisher={IEEE}
}

@inproceedings{williams2017information,
  title={Information theoretic MPC for model-based reinforcement learning},
  author={Williams, Grady and Wagener, Nolan and Goldfain, Brian and Drews, Paul and Rehg, James M and Boots, Byron and Theodorou, Evangelos A},
  booktitle={2017 IEEE international conference on robotics and automation (ICRA)},
  pages={1714--1721},
  year={2017},
  organization={IEEE}
}

@article{rutherford2010modelling,
  title={Modelling nonlinear vehicle dynamics with neural networks},
  author={Rutherford, Simon J and Cole, David J},
  journal={International journal of vehicle design},
  volume={53},
  number={4},
  pages={260--287},
  year={2010},
  publisher={Inderscience Publishers}
}

@INPROCEEDINGS{suminaka2025adaptable,
  author={Suminaka, Makoto and Dallas, James and Thompson, Michael and Soga, Masayuki and Kasai, Eiji and Subosits, John},
  booktitle={2025 American Control Conference (ACC)}, 
  title={Adaptable High-Speed Model Predictive Control for Autonomous Drifting: Koopman-Based Dynamics}, 
  year={2025},
  volume={},
  number={},
  pages={1057-1063},
  doi={10.23919/ACC63710.2025.11107568}
}

@inproceedings{djeumou2024one,
  title={One model to drift them all: Physics-informed conditional diffusion model for driving at the limits},
  author={Djeumou, Franck and Lew, Thomas Jonathan and Ding, Nan and Thompson, Michael and Suminaka, Makoto and Greiff, Marcus and Subosits, John},
  booktitle={8th Annual Conference on Robot Learning},
  year={2024}
}

@inproceedings{ding2024drifting,
  title={Drifting with unknown tires: Learning vehicle models online with neural networks and model predictive control},
  author={Ding, Nan and Thompson, Michael and Dallas, James and Goh, Jonathan YM and Subosits, John},
  booktitle={2024 IEEE Intelligent Vehicles Symposium (IV)},
  pages={2545--2552},
  year={2024},
  organization={IEEE}
}

@inproceedings{djeumou2023autonomous,
  title={Autonomous drifting with 3 minutes of data via learned tire models},
  author={Djeumou, Franck and Goh, Jonathan YM and Topcu, Ufuk and Balachandran, Avinash},
  booktitle={2023 IEEE International Conference on Robotics and Automation (ICRA)},
  pages={968--974},
  year={2023},
  organization={IEEE}
}

@article{nguyen2011model,
  title={Model learning for robot control: a survey},
  author={Nguyen-Tuong, Duy and Peters, Jan},
  journal={Cognitive processing},
  volume={12},
  number={4},
  pages={319--340},
  year={2011},
  publisher={Springer}
}

@article{brunke2022safe,
  title={Safe learning in robotics: From learning-based control to safe reinforcement learning},
  author={Brunke, Lukas and Greeff, Melissa and Hall, Adam W and Yuan, Zhaocong and Zhou, Siqi and Panerati, Jacopo and Schoellig, Angela P},
  journal={Annual Review of Control, Robotics, and Autonomous Systems},
  volume={5},
  number={1},
  pages={411--444},
  year={2022},
  publisher={Annual Reviews}
}

@ARTICLE{zhang2023tire,
  author={Zhang, Rongyun and Feng, Yongle and Shi, Peicheng and Zhao, Linfeng and Du, Yufeng and Liu, Yaming},
  journal={IEEE Transactions on Vehicular Technology}, 
  title={Tire-Road Friction Coefficient Estimation for Distributed Drive Electric Vehicles Using PMSM Sensorless Control}, 
  year={2023},
  volume={72},
  number={7},
  pages={8672-8685},
  doi={10.1109/TVT.2023.3248866}}

@INPROCEEDINGS{li2021dual,
  author={Li, Chenran and Liu, Yulong and Sun, Liting and Liu, Yahui and Tomizuka, Masayoshi and Zhan, Wei},
  booktitle={2021 IEEE International Intelligent Transportation Systems Conference (ITSC)}, 
  title={Dual Extended Kalman Filter Based State and Parameter Estimator for Model-Based Control in Autonomous Vehicles}, 
  year={2021},
  volume={},
  number={},
  pages={327-333},
  doi={10.1109/ITSC48978.2021.9564571}}

@ARTICLE{wang2021integrated,
  author={Wang, Yan and Lv, Chen and Yan, Yongjun and Peng, Pai and Wang, Faan and Xu, Liwei and Yin, Guodong},
  journal={IEEE Transactions on Industrial Electronics}, 
  title={An Integrated Scheme for Coefficient Estimation of Tire–Road Friction With Mass Parameter Mismatch Under Complex Driving Scenarios}, 
  year={2022},
  volume={69},
  number={12},
  pages={13337-13347},
  doi={10.1109/TIE.2021.3134072}}

@ARTICLE{qin2022lateral,
  author={Qin, Zhaobo and Chen, Liang and Hu, Manjiang and Chen, Xin},
  journal={IEEE Transactions on Vehicular Technology}, 
  title={A Lateral and Longitudinal Dynamics Control Framework of Autonomous Vehicles Based on Multi-Parameter Joint Estimation}, 
  year={2022},
  volume={71},
  number={6},
  pages={5837-5852},
  doi={10.1109/TVT.2022.3163507}}

@article{wenzel2006dual,
  title={Dual extended Kalman filter for vehicle state and parameter estimation},
  author={Wenzel, Thomas A and Burnham, KJ and Blundell, MV and Williams, RA},
  journal={Vehicle system dynamics},
  volume={44},
  number={2},
  pages={153--171},
  year={2006},
  publisher={Taylor \& Francis}
}

@article{xu2024combined,
  title={Combined-slip trajectory tracking and yaw stability control for 4WID autonomous vehicles based on effective cornering stiffness},
  author={Xu, Nan and Hu, Min and Jin, Lingge and Ding, Haitao and Huang, Yanjun},
  journal={IEEE Transactions on Intelligent Transportation Systems},
  year={2024},
  publisher={IEEE}
}

@article{guo2018tire,
  title={Tire side force characteristics with the coupling effect of vertical load and inflation pressure},
  author={Guo, Konghui and Chen, Ping and Xu, Nan and Yang, Chao and Li, Fei},
  journal={SAE International Journal of Vehicle Dynamics, Stability, and NVH},
  volume={3},
  number={10-03-01-0002},
  pages={19--30},
  year={2018}
}

@article{christ2021time,
  title={Time-optimal trajectory planning for a race car considering variable tyre-road friction coefficients},
  author={Christ, Fabian and Wischnewski, Alexander and Heilmeier, Alexander and Lohmann, Boris},
  journal={Vehicle system dynamics},
  volume={59},
  number={4},
  pages={588--612},
  year={2021},
  publisher={Taylor \& Francis}
}

@article{brown2017safe,
  title={Safe driving envelopes for path tracking in autonomous vehicles},
  author={Brown, Matthew and Funke, Joseph and Erlien, Stephen and Gerdes, J Christian},
  journal={Control Engineering Practice},
  volume={61},
  pages={307--316},
  year={2017},
  publisher={Elsevier}
}

@article{subosits2019racetrack,
  title={From the racetrack to the road: Real-time trajectory replanning for autonomous driving},
  author={Subosits, John K and Gerdes, J Christian},
  journal={IEEE Transactions on Intelligent Vehicles},
  volume={4},
  number={2},
  pages={309--320},
  year={2019},
  publisher={IEEE}
}

@article{falcone2007predictive,
  title={Predictive active steering control for autonomous vehicle systems},
  author={Falcone, Paolo and Borrelli, Francesco and Asgari, Jahan and Tseng, Hongtei Eric and Hrovat, Davor},
  journal={IEEE Transactions on control systems technology},
  volume={15},
  number={3},
  pages={566--580},
  year={2007},
  publisher={IEEE}
}

@article{brown2019coordinating,
  title={Coordinating tire forces to avoid obstacles using nonlinear model predictive control},
  author={Brown, Matthew and Gerdes, J Christian},
  journal={IEEE Transactions on Intelligent Vehicles},
  volume={5},
  number={1},
  pages={21--31},
  year={2019},
  publisher={IEEE}
}

@article{chen2020implementation,
  title={Implementation of MPC-based path tracking for autonomous vehicles considering three vehicle dynamics models with different fidelities},
  author={Chen, Shuping and Chen, Huiyan and Negrut, Dan},
  journal={Automotive Innovation},
  volume={3},
  number={4},
  pages={386--399},
  year={2020},
  publisher={Springer}
}

@article{frison2020hpipm,
  title={HPIPM: a high-performance quadratic programming framework for model predictive control},
  author={Frison, Gianluca and Diehl, Moritz},
  journal={IFAC-PapersOnLine},
  volume={53},
  number={2},
  pages={6563--6569},
  year={2020},
  publisher={Elsevier}
}

@article{kuutti2020survey,
  title={A survey of deep learning applications to autonomous vehicle control},
  author={Kuutti, Sampo and Bowden, Richard and Jin, Yaochu and Barber, Phil and Fallah, Saber},
  journal={IEEE Transactions on Intelligent Transportation Systems},
  volume={22},
  number={2},
  pages={712--733},
  year={2020},
  publisher={IEEE}
}

@article{spielberg2021neural,
  title={Neural network model predictive motion control applied to automated driving with unknown friction},
  author={Spielberg, Nathan A and Brown, Matthew and Gerdes, J Christian},
  journal={IEEE Transactions on Control Systems Technology},
  volume={30},
  number={5},
  pages={1934--1945},
  year={2021},
  publisher={IEEE}
}

@article{spielberg2019neural,
  title={Neural network vehicle models for high-performance automated driving},
  author={Spielberg, Nathan A and Brown, Matthew and Kapania, Nitin R and Kegelman, John C and Gerdes, J Christian},
  journal={Science robotics},
  volume={4},
  number={28},
  pages={eaaw1975},
  year={2019},
  publisher={American Association for the Advancement of Science}
}

@inproceedings{pan2017prediction,
  title={Prediction under uncertainty in sparse spectrum Gaussian processes with applications to filtering and control},
  author={Pan, Yunpeng and Yan, Xinyan and Theodorou, Evangelos A and Boots, Byron},
  booktitle={International Conference on Machine Learning},
  pages={2760--2768},
  year={2017},
  organization={PMLR}
}

@inproceedings{mesbah2022fusion,
  title={Fusion of machine learning and MPC under uncertainty: What advances are on the horizon?},
  author={Mesbah, Ali and Wabersich, Kim P and Schoellig, Angela P and Zeilinger, Melanie N and Lucia, Sergio and Badgwell, Thomas A and Paulson, Joel A},
  booktitle={2022 American Control Conference (ACC)},
  pages={342--357},
  year={2022},
  organization={IEEE}
}

@article{kabzan2020amz,
  title={AMZ driverless: The full autonomous racing system},
  author={Kabzan, Juraj and Valls, Miguel I and Reijgwart, Victor JF and Hendrikx, Hubertus FC and Ehmke, Claas and Prajapat, Manish and B{\"u}hler, Andreas and Gosala, Nikhil and Gupta, Mehak and Sivanesan, Ramya and others},
  journal={Journal of Field Robotics},
  volume={37},
  number={7},
  pages={1267--1294},
  year={2020},
  publisher={Wiley Online Library}
}

@article{snelson2005sparse,
  title={Sparse Gaussian processes using pseudo-inputs},
  author={Snelson, Edward and Ghahramani, Zoubin},
  journal={Advances in neural information processing systems},
  volume={18},
  year={2005}
}

@article{lazaro2010sparse,
  title={Sparse spectrum Gaussian process regression},
  author={L{\'a}zaro-Gredilla, Miguel and Quinonero-Candela, Joaquin and Rasmussen, Carl Edward and Figueiras-Vidal, An{\'\i}bal R},
  journal={The Journal of Machine Learning Research},
  volume={11},
  pages={1865--1881},
  year={2010},
  publisher={JMLR. org}
}

@article{nguyen2011incremental,
  title={Incremental online sparsification for model learning in real-time robot control},
  author={Nguyen-Tuong, Duy and Peters, Jan},
  journal={Neurocomputing},
  volume={74},
  number={11},
  pages={1859--1867},
  year={2011},
  publisher={Elsevier}
}

@book{pacejka2005tire,
  title={Tire and vehicle dynamics},
  author={Pacejka, Hans},
  year={2005},
  publisher={Elsevier}
}

@article{rossa2012bifurcation,
  title={Bifurcation analysis of an automobile model negotiating a curve},
  author={Rossa, Fabio Della and Mastinu, Giampiero and Piccardi, Carlo},
  journal={Vehicle system dynamics},
  volume={50},
  number={10},
  pages={1539--1562},
  year={2012},
  publisher={Taylor \& Francis}
}

@article{jiang2023fast,
  title={A Fast Kernel-Based Optimal Control Framework for Autonomous Driving},
  author={Jiang, Chunmao and Ding, Yi and Li, Zhiyuan and Sun, Chao},
  journal={IEEE Transactions on Control Systems Technology},
  volume={31},
  number={3},
  pages={1296--1307},
  year={2023},
  publisher={IEEE}
}

@inproceedings{hensman2015scalable,
  title={Scalable variational Gaussian process classification},
  author={Hensman, James and Matthews, Alexander and Ghahramani, Zoubin},
  booktitle={Artificial Intelligence and Statistics},
  pages={351--360},
  year={2015},
  organization={PMLR}
}

\end{document}